\documentclass[runningheads]{llncs}

\usepackage{eccv}

\usepackage{eccvabbrv}

\usepackage{graphicx}
\usepackage{booktabs}
\usepackage{natbib}
\setcitestyle{numbers,square,citesep={,},aysep={,},yysep={,}}
\usepackage{times}
\usepackage{epsfig}
\usepackage{graphicx}
\usepackage{amsmath}
\usepackage{amssymb}
\usepackage{graphicx}  
\usepackage{tikz}  
\usepackage[inline]{enumitem}
\usepackage{booktabs}
\usepackage{float}
\usepackage{algorithm, algpseudocodex}
\usepackage{nicefrac}
\usepackage{enumitem}
\usepackage{makecell}
\usepackage{anyfontsize}
\usepackage{caption}
\usepackage{mathtools}
\usepackage[bottom, multiple]{footmisc} 
\usepackage{tikz} 
\usepackage{subcaption}
\usepackage{xspace}
\usepackage[nolist,nohyperlinks]{acronym}
\usepackage[separate-uncertainty=true,detect-weight=true,detect-inline-weight=math]{siunitx}
\usepackage{pgfplots}
\usepackage{nicefrac}
\usepackage[leftcaption]{sidecap}
\usepackage{wrapfig}

\usepackage[accsupp]{axessibility}  

\usepackage[pagebackref,breaklinks,colorlinks,citecolor=eccvblue]{hyperref}

\usepackage{orcidlink}

\newcommand{\methodname}{RadEdit\xspace}
\newcommand{\covid}{COVID-19\xspace}
\newcommand{\biovil}{\mbox{BioViL-T}\xspace}

\newcommand{\prompt}[1]{\textit{`#1'}}

\definecolor{mygrey}{HTML}{212121}
\definecolor{figurepurple}{HTML}{7146bb}
\definecolor{figureblue}{HTML}{0078d4}
\definecolor{figureorange}{HTML}{ff7c5a}
\definecolor{figuregreen}{HTML}{65ce62}
\definecolor{myblue}{HTML}{467cb2}
\definecolor{myred}{HTML}{c83129}

\begin{document}

\sisetup{mode=match}

\begin{acronym}
    \acro{AHD}{average Hausdorff distance}
    \acro{AUROC}{area under the receiver operating characteristic curve}
    \acro{CFG}{classifier-free guidance}
    \acro{DDIM}{denoising diffusion implicit model}
    \acro{DDPM}{denoising diffusion probabilistic model}
    \acro{GAN}{generative adversarial network}
    \acro{LLM}{large language model}
    \acro{VAE}{variational autoencoder}
\end{acronym}

\title{\methodname: stress-testing biomedical vision models via diffusion image editing} 

\titlerunning{RadEdit}

\author{Fernando P\'erez-Garc\'ia$^{\star,}$\inst{1} \and
Sam Bond-Taylor$^{\star,}$\inst{1} 
\and Pedro P. Sanchez$^{+,}$\inst{2} 
\and Boris van Breugel$^{+,}$\inst{3} 
\and Daniel C. Castro\inst{1}
\and Harshita Sharma\inst{1}
\and Valentina Salvatelli\inst{1}
\and Maria T.A. Wetscherek\inst{1}
\and Hannah Richardson\inst{1}
\and Matthew P. Lungren\inst{1, 4, 5}
\and Aditya Nori\inst{1}
\and Javier Alvarez-Valle\inst{1} 
\and Ozan Oktay $^{\dagger,}$\inst{1}
\and Maximilian Ilse $^{\dagger,}$\inst{1}}

\authorrunning{F.~P\'erez-Garc\'ia and S.~Bond-Taylor et al.}

\institute{Microsoft Health Futures
\and University of Edinburgh
\and University of Cambridge
\and University if California
\and Stanford University
\\$^\star$ Shared first author
\\$^+$ Work done at Microsoft Health Futures \\$^\dagger$ Shared last author}

\maketitle

\begin{abstract}\vspace{-0.6em}
  Biomedical imaging datasets are often small and biased, meaning that real-world performance of predictive models can be substantially lower than expected from internal testing. This work proposes using generative image editing to simulate dataset shifts and diagnose failure modes of biomedical vision models; this can be used in advance of deployment to assess readiness, potentially reducing cost and patient harm. Existing editing methods can produce undesirable changes, with spurious correlations learned due to the co-occurrence of disease and treatment interventions, limiting practical applicability. To address this, we train a text-to-image diffusion model on multiple chest X-ray datasets and introduce a new editing method, \methodname, that uses multiple image masks, if present, to constrain changes and ensure consistency in the edited images, minimising bias. We consider three types of dataset shifts: acquisition shift, manifestation shift, and population shift, and demonstrate that our approach can diagnose failures and quantify model robustness without additional data collection, complementing more qualitative tools for explainable AI.
  \keywords{Image editing \and diffusion models \and biomedical imaging}
\end{abstract}

\section{Introduction}\vspace{-0.3em}

\begin{figure}[t]
    \centering
    \begin{minipage}{0.42\linewidth}
        \caption{\textbf{Stress-testing models by simulating dataset shifts via image editing.}\\ \textit{Top}: editing out \covid features results in false positives since the classifier relies on acquisition differences, e.g., radiographic markers (white arrow). \\ \textit{Middle}: editing out a pneumothorax (PTX) results in false positives since the classifier instead detects chest drains. \\ \textit{Bottom}: editing abnormalities into lungs causes a lung segmentation model to mislabel (\textcolor{myblue}{blue}: ground-truth segmentation; \textcolor{myred}{red}: model prediction).}
    \end{minipage}
    \hfill
    \begin{minipage}{0.55\linewidth}
        \resizebox{\linewidth}{!}{
        \begin{picture}(231,168)
            \put(-2,0){\includegraphics[width=8.3cm]{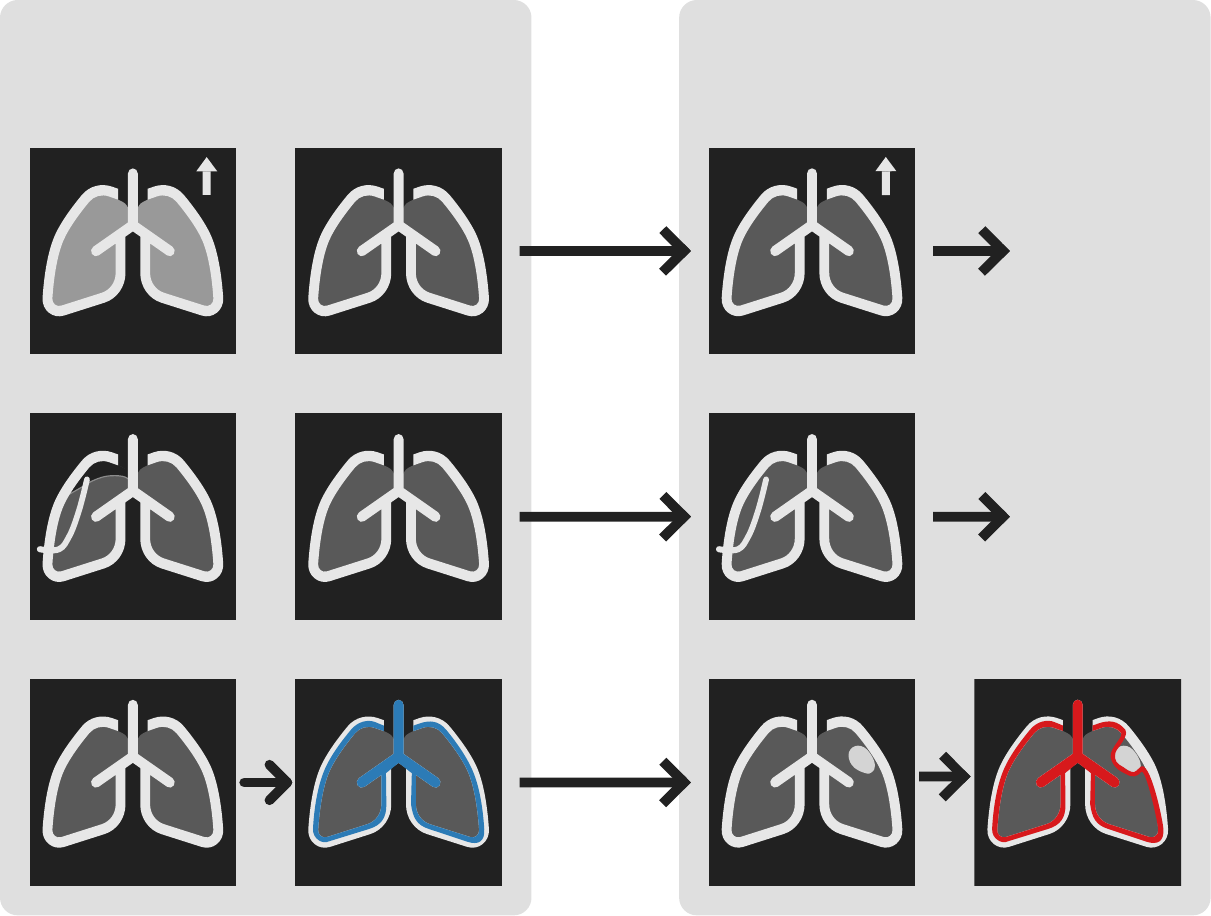}}
            \put(5,164){\textcolor{mygrey}{\fontfamily{phv}\fontsize{8}{8}\selectfont  Biased training datasets}}
            \put(136,164){\textcolor{mygrey}{\fontfamily{phv}\fontsize{8}{8}\selectfont  Edited stress-testing set}}
            \put(10,151){\textcolor{mygrey}{\fontfamily{phv}\fontsize{6}{6}\selectfont  \covid}}
            \put(57,151){\textcolor{mygrey}{\fontfamily{phv}\fontsize{6}{6}\selectfont No \covid}}
            \put(17.5,99.3){\textcolor{mygrey}{\fontfamily{phv}\fontsize{6}{6}\selectfont  PTX}}
            \put(64,99.3){\textcolor{mygrey}{\fontfamily{phv}\fontsize{6}{6}\selectfont  No PTX}}
            \put(13,48){\textcolor{mygrey}{\fontfamily{phv}\fontsize{6}{6}\selectfont  Healthy}}
            \put(57.5,48){\textcolor{mygrey}{\fontfamily{phv}\fontsize{6}{6}\selectfont  Segmentation}}
            \put(101.3,132){\textcolor{mygrey}{\fontfamily{phv}\fontsize{5}{5}\selectfont  Acquisition}}
            \put(109,123){\textcolor{mygrey}{\fontfamily{phv}\fontsize{5}{5}\selectfont  shift}}
            \put(100.5,83){\textcolor{mygrey}{\fontfamily{phv}\fontsize{5}{5}\selectfont  Manifestation}}
            \put(109,71.5){\textcolor{mygrey}{\fontfamily{phv}\fontsize{5}{5}\selectfont  shift}}
            \put(102,29){\textcolor{mygrey}{\fontfamily{phv}\fontsize{5}{5}\selectfont Population}}
            \put(109,20){\textcolor{mygrey}{\fontfamily{phv}\fontsize{5}{5}\selectfont  shift}}
            \put(138,151){\textcolor{mygrey}{\fontfamily{phv}\fontsize{6}{6}\selectfont  No COVID-19}}
            \put(145,99.3){\textcolor{mygrey}{\fontfamily{phv}\fontsize{6}{6}\selectfont  No PTX}}
            \put(140,48){\textcolor{mygrey}{\fontfamily{phv}\fontsize{6}{6}\selectfont  Abnormality}}
            \put(201,80.5){\textcolor{mygrey}{\fontfamily{phv}\fontsize{8}{8}\selectfont  False}}
            \put(198,71.5){\textcolor{mygrey}{\fontfamily{phv}\fontsize{8}{8}\selectfont  positive}}
            \put(201,130){\textcolor{mygrey}{\fontfamily{phv}\fontsize{8}{8}\selectfont  False}}
            \put(198,122){\textcolor{mygrey}{\fontfamily{phv}\fontsize{8}{8}\selectfont  positive}}
            \put(192,48){\textcolor{mygrey}{\fontfamily{phv}\fontsize{6}{6}\selectfont  Mislabelled}}
            
        \end{picture}
        }
    \end{minipage}
    \vspace{-1em}
\end{figure}

Developing accurate and robust models for biomedical image analysis requires large and diverse datasets that are often difficult to obtain due to ethical, legal, geographical, and financial constraints \cite{lee_medical_2017}. This leads to biased training datasets that affect the performance of trained models and generalisation to real-world scenarios \cite{rueckel_pneumothorax_2021, larrazabal_gender_2020}. Such data mismatch may arise from genuine differences in upstream data acquisition as well as from the selection criteria for dataset creation, which materialise as various forms of dataset shifts (population, acquisition, annotation, prevalence, manifestation) \cite{castro_2020_causality}.

Biomedical vision models, when put into real-world use, can be unhelpful or potentially even harmful to patients if they are affected by dataset shifts, leading to missed diagnoses \citep{heaven_hundreds_2021, von_data_2021, wynants_prediction_2020, roberts_common_2021}. For example, the \covid pandemic led to hundreds of detection tools being developed, with some put into use in hospitals; yet \citet{roberts_common_2021} found that \textit{``none of the models identified are of potential clinical use due to methodological flaws and/or underlying biases.''} It is therefore crucial to properly assess models for bias, prior to real-world use.

Recent deep generative models have made remarkable improvements in terms of sample quality, diversity, and steerability \cite{rombach_high-resolution_2022, muller_2022_diffusion, kang_2023_scaling, ho_2022_cascaded}. These models have been shown to generalise to out-of-distribution domains \cite{li_imagenet-e_2023, barbano_2023_steerable, jaini_2023_intriguing, garipov_compositional_2023}, opening up avenues for new applications. One such application is generating synthetic data for stress-testing models \cite{prabhu_lance_2023, li_imagenet-e_2023, van_breugel_can_2023}. This involves creating data that is realistic, yet can represent settings, domains, or populations that do not appear (enough) in the real training/test data. 

In this work, we investigate how deep generative models can be used for stress-testing biomedical imaging models.
We consider three dataset shift scenarios:
\begin{enumerate}[topsep=4pt]
    \item \textbf{Acquisition shift:} classifying \covid cases when the positive and negative cases were acquired at different hospitals (\cref{sec:covid_exp}).\label{t1}
    \item \textbf{Manifestation shift:} detecting if pneumothorax\footnote{\label{foot:terms} We provide descriptions of the medical terms used throughout the paper in \cref{app:terms}} was resolved when chest drains (inserted to treat pneumothorax) are present (\cref{sec:ptx_exp}).\label{t2}
    \item \textbf{Population shift:} segmenting lungs in the presence of abnormalities rarely or never seen in the training dataset (\cref{sec:seg_exp}).\label{t3}
\end{enumerate}
For each of these scenarios, we simulate dataset shifts, producing stress-test sets which can occur in the real world but do not appear or are underrepresented in the original training/test sets. Following prior work, these test sets are synthesised using generative image editing, which unlike generating images from scratch, only minimally modifies the images, hence, better retains fidelity and diversity \cite{prabhu_lance_2023, li_imagenet-e_2023}. For the above scenarios, we use generative editing to 
\begin{enumerate*}
    \item remove only \covid features while keeping visual indicators of the different hospitals; 
    \item remove only pneumothorax while keeping the chest drain; and
    \item add abnormalities that occlude lung structures in the image.
\end{enumerate*}

We train a generative diffusion model \cite{ho_denoising_2020,rombach_high-resolution_2022} on a large number of chest X-rays from a variety of biomedical imaging datasets (\cref{sec:foundation_diff_model}). The diversity within this training data enables us to add and remove a wide variety of pathologies and support devices when editing.
Despite the diversity within these datasets, substantial biases are still present, some of which are learned by the generative model. As a result, when using diffusion models for image editing, correlated features may also be modified. For example, in Scenario \labelcref{t2}, removing the pneumothorax might also remove the chest drains as both features typically co-occur in datasets \cite{Rueckel2020}, since chest drains are used to treat pneumothorax. Furthermore, when editing only within editing masks, artefacts often appear at the border of the masks. Lastly, artefacts occur when editing images outside of the training dataset domain of the diffusion model used for editing. To overcome these challenges, we propose using multiple masks to break existing correlations. This involves defining which regions must change, and explicitly forcing correlated regions to remain unchanged. In addition, we allow the area outside of the masks to be modified by the diffusion model to ensure image consistency. Since our proposed editing method, which we call \methodname, leads to only minimal overall changes of chest X-rays, we are able to generate synthetic datasets that can be used to stress-test segmentation models (Scenario \labelcref{t3}), which, to the best of our knowledge, we are the first to demonstrate. 


In summary, our contributions are as follows: 
\begin{itemize}[topsep=1pt]
    \item We introduce a novel editing approach that reduces the presence of artefacts in edited images and simplifies prompt construction compared to prior work \cite{couairon_diffedit_2022, prabhu_lance_2023}.
    \item Our editing approach allows us to construct synthetic datasets with specific data shifts by performing zero-shot edits on datasets/abnormalities not seen in training. 
    \item We conduct a broad set of experiments using these synthetic datasets to stress-test and expose biases in biomedical classification and, for the first time, segmentation models, introducing a new use case of synthetic data into the medical setting.
\end{itemize}

\vspace{-0.3em}
\section{Related work}\vspace{-0.2em}
\vspace{-0.15em}
\subsection{Generative image editing}\vspace{-0.1em}
With advances in deep generative modelling, several approaches to image editing have emerged. Many of these early approaches use compressed latent manipulation \cite{dosovitskiy_learning_2015, radford_unsupervised_2015, shen_interpreting_2020, upchurch_deep_2017} where fine-grained edits are difficult to achieve and can result in unwanted changes. More recently, the unparalleled flexibility of diffusion models, together with advances in plain text conditioning, have opened up new avenues for editing techniques. 

Here, we describe some notable diffusion editing methods.
SDEdit \cite{meng_sdedit_2022} shows that diffusion models trained solely on real images can be used to generate images from sketches by perturbing sketches with noise, then running the reverse diffusion process.
Palette \cite{saharia_palette_2022} is an image-to-image diffusion model that can be used for inpainting by filling a masked region with noise and denoising that region.
Blended diffusion \cite{avrahami_blended_2022, avrahami_blended_2023} uses masks with CLIP \cite{radford_learning_2021} conditioning to guide local edits.
Multiple works show that injecting U-Net activations, obtained by encoding the original image into the generation process, makes the global structure of the source and edited images closely match \cite{hertz_prompt_2022, tumanyan_plug_2023}.
DiffEdit \cite{couairon_diffedit_2022} uses text prompts to determine the appropriate region to edit.
\citet{mokady_null_2023} improve diffusion inversion quality by optimising the diffusion trajectory. 

Crucially, in the works which use masks for editing, a single type of mask is always used to define the region of interest. In this work, we argue that a second type of mask is required to avoid the loss of features caused by spurious correlations. As better editing approaches are developed, this requirement should be kept in mind.

\vspace{-0.15em}
\subsection{Stress-testing}\vspace{-0.1em}
Several approaches have used non-deep-generative-model methods to stress-test networks.
\citet{hendrycks_benchmarking_2018} evaluate classification models' robustness to corruptions such as blurring, Gaussian noise, and JPEG artefacts.
\citet{sakaridis_semantic_2018} stress-test a segmentation model for roads by using an optical model to add synthetic fog to scenes.
\citet{koh_2021_wilds} collate a dataset presenting various distribution shifts.

More recent models have made use of conditional generative models to simulate shifts. \citet{prabhu_lance_2023} propose LANCE, which stress-tests classification models by using diffusion-based image editing to modify image subjects via caption editing with a \ac{LLM}; \citet{kattakinda_invariant_2022} do similar, but instead modify the background. 
\citet{li_imagenet-e_2023} use diffusion models with a single subject mask to separately edit backgrounds and subjects.
\Citet{van_breugel_can_2023} use generative adversarial networks to simulate distribution shifts on tabular data.
This line of research is partially related to adversarial attacks \cite{goodfellow_explaining_2015}, where the focus is on minimally modifying images such that they are visually indistinguishable to a human, but the attacked model fails.

\vspace{-0.1em}
\subsection{Biomedical imaging counterfactuals}\vspace{-0.1em}
Generative models have also been applied to biomedical counterfactual generation.
\citet{reinhold_structural_2021} manipulate causes of multiple sclerosis in brain MRI with deep structural causal models \citep{pawlowski_dscm_2020}.
\citet{sanchez_healthy_2022} and \citet{fontanella_diffusion_2023} use editing to remove pathologies for abnormality detection.
\citet{ktena_generative_2023} generate out-of-distribution samples to improve classifier performance.
\citet{gu_biomedjourney_2023} train a diffusion model to model disease progression by conditioning on a prior X-ray and text progression description.
Unlike our approach, these methods do not use masks to enforce which regions may or may not be edited, meaning that spurious correlations might affect edits. Additionally, these methods use synthetic data to augment and improve model performance whereas we focus on using synthetic medical data for stress-testing.

\vspace{-0.3em}
\section{Preliminaries}\vspace{-0.2em}
In this section, we introduce background context for stress-testing biomedical imaging models: failure modes of biomedical imaging models caused by different dataset shifts; diffusion models as versatile generative models; and diffusion-based image editing.

\vspace{-0.1em}
\subsection{Dataset shifts}\vspace{-0.1em}
Dataset shift refers to a discrepancy between the training and test data distributions due to external factors \cite{castro_2020_causality,jones_no_2023}. Such shifts are regularly observed in machine learning for biomedical imaging, often due to data scarcity. For example, collected training datasets might consist primarily of healthy patients. However, when the model is used in practice after training, there could be a shift towards unhealthy patients. A taxonomy of different types of dataset shifts in the context of biomedical imaging was developed by \citet{castro_2020_causality}. In this paper, we consider three dataset shifts of particular interest.

\vspace{-0.33em}
\paragraph{Acquisition shift} 
results from the use of different scanners (manufacturer, hardware, and software) or imaging protocols as often encountered when using data from multiple cohorts. These changes affect factors such as image resolution, contrast, patient positioning, and image markings. 
\vspace{-0.33em}

\paragraph{Manifestation shift} 
results from the way the prediction targets physically manifest in anatomy changes between domains. For example, training datasets could consist of more severe pathological cases than observed in practice, or a pathology may co-occur with different visual features, e.g., support devices. 

\paragraph{Population shift} 
results from differences in intrinsic characteristics of the populations under study, changing the anatomical appearance distribution. This definition encompasses examples such as age, sex, ethnicity, and comorbidities, but also abnormalities such as pleural effusion and support devices. In contrast to manifestation shift, the shift in anatomical appearance is not affected by prediction targets.
\vspace{-0.33em}

\vspace{-0.2em}
\subsection{Diffusion models}\vspace{-0.1em}
\Acp{DDPM} \cite{ho_denoising_2020, sohl_deep_2015} are a versatile and effective class of generative models that enable sampling from the data distribution by learning to denoise samples corrupted with Gaussian noise.
\Acp{DDPM} are formed by defining a forward time process that gradually adds noise to data points $x_0$ through the recursion 
\vspace{-0.2em}\begin{equation}
    x_t = \sqrt{1-\beta_t} x_{t-1} + \sqrt{\beta_t} \epsilon_t, \quad t=1,\dots,T, \quad
    \text{s.t. } x_t = \sqrt{\bar{\alpha}_t} x_0 + \sqrt{1 - \bar{\alpha}_t} \bar{\epsilon}_t \,, \label{eq:forward_process}
\end{equation}\vspace{-0.2em}
where $\epsilon_{1:T}, \bar{\epsilon}_{1:T} \sim \mathcal{N}(0, I)$, $\beta_{1:T}$ is a predefined noise schedule that determines how quickly to corrupt the data and ensures that $x_T$ contains little to no information about $x_0$, and $\bar{\alpha}_t=\prod_{s=1}^t (1-\beta_s)$. To form a generative model, the process is reversed in time, gradually transforming Gaussian noise into samples from the learned distribution. While the exact reversal is intractable, a variational approximation is defined by \citep{song_denoising_2022}:
%
\vspace*{-1em}\begin{gather}
    x_{t-1}\!=\!\hat{\mu}_t(x_t, f_\theta(x_t, t, c)) + \sigma_t z_t, \\
    \hat{\mu}_t(x_t, \epsilon_t) \! = \! \sqrt{\bar{\alpha}_{t-1}}\frac{x_t \! - \! \sqrt{1{-}\bar{\alpha}_t}\epsilon_t}{\sqrt{\bar{\alpha}_{t}}}  + \sqrt{1\!-\!\bar{\alpha}_{t-1} \! - \! \sigma^2_t} \epsilon_t,
\end{gather}
%
where $c$ is a conditioning signal such as a text description, $f_\theta(x_t, t, c)$ is a learned approximation of the noise $\bar{\epsilon}_t$ that corrupted the image $x_0$ to obtain $x_t$, $z_{1:T} \sim \mathcal{N}(0, I)$, and $\sigma_{1:T}$ controls how much noise is introduced. The process is Markovian and known as a DDPM \cite{ho_denoising_2020} when $\sigma_t\!=\!\sqrt{\nicefrac{(1-\bar{\alpha}_{t-1})}{(1-\bar{\alpha}_t)}}\sqrt{1-\nicefrac{\bar{\alpha}_t}{\bar{\alpha}_{t-1}}}$, while for $\sigma_t\!=\!0$ the process is deterministic and is called a \ac{DDIM} \cite{song_denoising_2022}.

\vspace{-0.2em}
\subsection{Image editing}\vspace{-0.1em}
\label{sec:image-editing}
The deterministic nature of \ac{DDIM}s leads to samples having a one-to-one correspondence with latent vectors $x_T$. As a result, we can deterministically map data points to latent vectors by running the \ac{DDIM} generative process in reverse  \cite{song_denoising_2022}, called \ac{DDIM} inversion. Several approaches \cite{couairon_diffedit_2022, meng_sdedit_2022} have shown that images can be edited by running the reverse diffusion process augmented by the latent vectors and a modified prompt $c$.

However, editing with \ac{DDIM} inversion can lead to undesired artefacts in the edited images.
For example, structures unrelated to the desired edit may also change shape, size, or location.
To address this, \citet{huberman-spiegelglas_edit_2023} propose \ac{DDPM} inversion, which better retains structure when editing. Here, the original forward process defined in \cref{eq:forward_process} is adapted, replacing the correlated vectors $\bar{\epsilon}_{1:T}$ with statistically independent vectors $\tilde{\epsilon}_{1:T}$ (\cref{alg:1}). These noise vectors are then used in the generative process, retaining the structure of the original image better than \ac{DDIM} inversion.

\newcommand{\ddpminv}{\textsc{DdpmInversion}}
\algnewcommand{\LineComment}[1]{\State $\color{gray}\triangleright$ \textit{\color{gray} #1}}
\begin{figure}[t]
\vspace{-2.7em}
\hfill
\begin{minipage}{0.45\textwidth}
\begin{algorithm}[H]
    \captionsetup{font=small}
    \caption{\ac{DDPM} inversion \citep{huberman-spiegelglas_edit_2023}} 
    \begin{algorithmic}
    \Require{image $x_0$, inversion prompt $c_\text{inv}$, diffusion model $f_\theta$}
        \LineComment{Sample statistically independent $\tilde{\epsilon}_t$}
        \For{$t \gets 1$ to T}
            \State {$\tilde{\epsilon}_t \sim \mathcal{N}(0, I)$}
            \State {$\hat{x}_t$ $\gets$ $\sqrt{\bar{\alpha}_t}x_0 + \sqrt{1-\bar{\alpha}_t}\tilde{\epsilon}_t$}
        \EndFor
        \LineComment{Isolate $z_t$ from series $\hat{x}_{1:T}$} 
        \For{$t \gets T$ to 1}
        \State {$\epsilon_t$ $\gets$ $f_\theta(\hat{x}_t, t, c_\text{inv})$}
            \State {$z_t$ $\gets$ $(\hat{x}_{t-1} - \hat{\mu}_t(\hat{x}_t, \epsilon_t))/\sigma_t$}
            \LineComment{Avoid error accumulation}
            \State {$\hat{x}_{t-1}$ $\gets$ $\hat{\mu}_t(\hat{x}_t, \epsilon_t) + \sigma_t z_t$}
        \EndFor
        \State \Return $(\hat{x}_{1:T}, z_{1:T})$
    \end{algorithmic}
    \label{alg:1}
\end{algorithm}
\end{minipage}
\hfill
\begin{minipage}{0.49\textwidth}
\begin{algorithm}[H]
    \captionsetup{font=small}
    \caption{DiffEdit \citep{couairon_diffedit_2022} w/ \ac{DDPM} inversion}
    \begin{algorithmic}
    \Require{image $x_0$, inversion prompt $c_\text{inv}$, edit prompt $c$, edit mask $m_\text{edit}$, \acs{CFG} weight $w$, diffusion model $f_\theta$}
        \State {$(\hat{x}_{1:T}, z_{1:T})$ $\gets$ {\ddpminv($x_0, c_\text{inv}$)}}
        \State {$x_T$ $\gets$ {$\hat x_T$}}
        \For{$t \gets T$ to 1}
            \State {$\epsilon_{\text{cond}, t}$ $\gets$ {$f_\theta(x_t, t, c)$}}
            \State {$\epsilon_{\text{uncond}, t}$ $\gets$ {$f_\theta(x_t, t, c = \emptyset)$}}
            \LineComment{\acf{CFG}}
            \State {$\epsilon_t$ $\gets$ {$\epsilon_{\text{uncond}, t} + w (\epsilon_{\text{cond}, t} - \epsilon_{\text{uncond}, t})$}}
            \State{$x_{t-1}$ $\gets$ {$\hat{\mu}_t(x_t, \epsilon_t) + \sigma_t z_t$}}
            \State{$x_{t-1}$ $\gets$ {$m_\text{edit} \odot x_{t-1} + (1 - m_\text{edit}) \odot \hat{x}_{t-1}$}}
        \EndFor
        \State \Return edited version of $x_0$
    \end{algorithmic}
    \label{alg:2}
\end{algorithm}
\end{minipage}
\hfill
\vspace{-0.1em}
\end{figure}

\section{Method}\vspace{-0.3em}
Our objective is to create synthetic test data through image editing that simulates specific data shifts, to rigorously evaluate biomedical imaging models. This synthetic data is used to predict model robustness, eliminating need for additional real-world test data.

\vspace{-0.2em}
\subsection{Limitations of existing editing methods}\vspace{-0.1em}
Recent advancements in diffusion modelling have drastically improved image editing. However, two prevalent approaches, LANCE \cite{prabhu_lance_2023} and DiffEdit \cite{couairon_diffedit_2022}, produce artefacts in medical images, making them unsuitable for stress-testing biomedical vision models.

LANCE only uses a global prompt (no mask) for image editing. While effective in the natural image domain, it leads to artefacts in the biomedical domain. For example in \cref{sec:seg_exp}, we add pathologies and support devices to images of healthy lungs to stress-test lung segmentation models. Since we want to use the original lung mask in combination with the edited image for testing, we need to ensure that the position and shape of the lung borders are not altered during editing. In \cref{fig:ablation}, we show that LANCE changes the position and shape of the lung border thus the edited images become unsuitable for stress-testing segmentation models. In addition, we find that LANCE potentially removes support devices when prompted to remove pathologies, which is a direct effect of the correlations in the datasets used to train the diffusion model in \cref{sec:foundation_diff_model}, making LANCE unsuited for testing the robustness of biomedical vision models to manifestation shift, see \cref{app:artefacts_lance} for an in-depth analysis.

DiffEdit (\cref{alg:2}) addresses these issues by editing only inside an automatically predicted mask $m_\text{edit}$. However, its automatic mask prediction often mismatches the manually annotated ground-truth, especially for small and complex abnormalities like pneumothorax\footref{foot:terms} (\cref{sec:quant_ablation}). Moreover, spurious correlations learned by the diffusion model can lead to the inclusion of support devices in the automatically predicted masks. Furthermore, even when relying on manually annotated masks, DiffEdit can introduce sharp discrepancies at mask boundaries, leading to unrealistic artefacts, such as when adding consolidation that should partially occlude the lung border (\cref{subfig:consolidation-diffedit}). Therefore, DiffEdit is also unsuitable for the segmentation experiments in \cref{sec:seg_exp}. 

\subsection{Improved editing with \methodname}\vspace{-0.2em}
To address the issues outlined in the previous section, we propose \methodname: by introducing `keep' and `edit' masks into the editing process, \methodname explicitly specifies which areas must remain unchanged (keep) and which should be actively modified based on the conditioning signal (edit). Crucially, these masks need not be mutually exclusive, allowing changes in the unmasked regions to ensure global consistency. Using masks, we assume that spurious correlations are mostly non-overlapping \citep{locatello_object_2020}.

\methodname is detailed in \cref{alg:3}, where a number of key properties make \methodname more suitable for biomedical image editing than prior editing methods. Firstly, since we aim to edit only within the edit mask $m_\text{edit}$, \acf{CFG} \cite{ho_classifier-free_2022} is used only within this region, with high guidance values (following \cite{huberman-spiegelglas_edit_2023}, we use a value of 15) ensuring that pathologies are completely removed without drastically changing the rest of the image. This approach also simplifies choosing a prompt for editing since we do not have to take into account the effect of the prompt on the rest of the image. Secondly, we allow the area outside $m_\text{edit}$ to be modified via unconditional generation to ensure image consistency.
Lastly, from the edited $x_{t-1}$, any changes made within the region of the keep mask $m_\text{keep}$ are reverted, ensuring that this region remains the same. Note that instead of initiating our generating process from pure noise we set $x_T = \hat{x}_T$, where $\hat{x}_T$ is the last output of the \ac{DDPM} inversion.

\begin{algorithm}[t] 
    \captionsetup{font=small}
    \caption{\methodname (ours) uses multiple masks to decouple spurious correlations}
    \begin{algorithmic}
    \Require{original image $x_0$, inversion prompt $c_\text{inv}$, editing prompt $c$, edit mask $m_\text{edit}$, keep mask $m_\text{keep}$, \ac{CFG} weight $w$, diffusion model $f_\theta$}
        
        \State {$(\hat{x}_{1:T}, z_{1:T})$ $\gets$ {\ddpminv($x_0, c_\text{inv}$)}}
        \State {$x_T$ $\gets$ {$\hat x_T$}}
        \For{$t \gets T$ to 1}
            \State {$\epsilon_{\text{cond}, t}$ $\gets$ {$f_\theta(x_t, t, c)$}} \Comment{Predict conditional noise}
            \State {$\epsilon_{\text{uncond}, t}$ $\gets$ {$f_\theta(x_t, t, c = \emptyset)$}} \Comment{Predict unconditional noise}
            \State {$\epsilon_t$ $\gets$ {$\epsilon_{\text{uncond}, t} + w (\epsilon_{\text{cond}, t} - \epsilon_{\text{uncond}, t})$}} \Comment{Combine noise predictions with \ac{CFG}}
            \State{$\epsilon_t$ $\gets$ {$m_\text{edit} \odot \epsilon_t + (1 - m_\text{edit}) \odot \epsilon_{\text{uncond}, t}$}} \Comment{Use \ac{CFG} only within $m_\text{edit}$}
            \State{$x_{t-1}$ $\gets$ {$\hat{\mu}_t(x_t, \epsilon_t) + \sigma_t z_t$}} \Comment{Move to next time step}
            \State{$x_{t-1}$ $\gets$ {$m_\text{keep} \odot \hat{x}_{t-1} + (1 - m_\text{keep}) \odot x_{t-1}$}} \Comment{Undo edits within $m_\text{keep}$}
        \EndFor
        \State \Return edited version of $x_0$
    \end{algorithmic}
    \label{alg:3}
\end{algorithm}

In Fig. \ref{subfig:pneumothorax-ours},  \ref{subfig:consolidation-ours}, we show that \methodname enables artefact-free editing while preserving structures of interest. Because the anatomical layout remains intact after editing, masks still correspond to the same structures, therefore the same masks can be reused to stress-test segmentation models (\cref{sec:seg_exp}).
In practice, we use a latent diffusion model \cite{rombach_high-resolution_2022}, therefore all operations in \cref{alg:3} are performed in the latent space of a \ac{VAE} \cite{rombach_high-resolution_2022}; this does not limit the generality of the approach. 

\vspace{-0.3em}
\subsection{Using synthetic images to uncover bias}\vspace{-0.2em}
Despite advancements in biomedical computer vision, recent studies have shown that bias in training and test data can lead to unrealistically high performance of machine learning models on the test set \citep{rueckel_pneumothorax_2021,degrave_ai_2021}. In our experiments, we use \methodname to create high quality synthetic test datasets that realistically capture specific dataset shifts, allowing us to quantify the robustness of models to these dataset shifts. By using masks, we can precisely edit the original training data to represent either acquisition shift, manifestation shift, or population shift \citep{castro_2020_causality} (\crefrange{sec:covid_exp}{sec:seg_exp}).
These synthetic test sets are used to stress-test (potentially biased) biomedical vision models by comparing performance to the real (biased) test set; a significant drop in performance indicates that the vision model is not robust to the dataset shift that can occur in clinical settings.
This serves as a complementary tool to visual explainable AI tools like Grad-CAM \cite{selvaraju_grad_2017} and saliency maps \cite{simonyan_deep_2013, adebayo_sanity_2018}, which offer qualitative insight into the robustness of models.

\vspace{-0.2em}
\subsection{\biovil editing score}\vspace{-0.2em}
\label{sec:clip-filtering}

\newcommand{\Ireal}{I_{\text{real}}}
\newcommand{\Treal}{T_{\text{real}}}
\newcommand{\Iedit}{I_{\text{edit}}}
\newcommand{\Tedit}{T_{\text{edit}}}

Since generative models result in samples of varying quality, poor-quality samples can be filtered out using image--text alignment scores, which quantitatively assess how closely related image--text pairs are via a pre-trained model that embeds similar images and text to nearby vectors \cite{azadi_dimscriminator_2018, razavi_generating_2019, radford_learning_2021, fernandez_privacy_2023}. For image editing, we instead assess how similar the change in text and image embeddings are after editing: for a real image--text pair $(\Ireal, \Treal)$, edited image--text pair $(\Iedit, \Tedit)$, image encoder $E_I$, and text encoder $E_T$, the editing score is defined based on directional similarity \citep{gal_stylegannada_2022}:
\vspace{-0.6em}
\begin{equation}
    \begin{aligned} 
        S_{\text{\biovil}} = \frac{\Delta I \cdot \Delta T}{\|\Delta I\| \|\Delta T\|} \,, \quad \text{where} \quad
    \end{aligned}
    \begin{aligned}
        \Delta I &= E_I(\Iedit) - E_I(\Ireal) \,,\,\text{and} \\
        \Delta T &= E_T(\Tedit) - E_T(\Treal)\,.
    \end{aligned}
\end{equation}
Given the focus on biomedical data, we use the \biovil~\cite{bannur_learning_2023} image and text encoders: domain-specific vision--language models trained to analyse chest X-rays and radiology reports, therefore well suited to measure changes in the edited image, such as removed pathologies. Following \citet{prabhu_lance_2023}, we discard images with $S_{\text{\biovil}} < 0.2$. This is not only effective for filtering out poor quality edits but is also able to detect whether the original image $\Ireal$ does not match the original text description $\Treal$ well. 

 \vspace{-0.2em}
 \section{Experiments} 
\label{sec:experiments}
\subsection{Diffusion model}\label{sec:foundation_diff_model} \vspace{-0.2em}
Our editing method is heavily dependent on a latent diffusion model \cite{rombach_high-resolution_2022} that can generate realistic chest X-rays. We use the \ac{VAE} \cite{kingma_auto-encoding_2022, higgins_beta_2016} of SDXL \cite{podell_sdxl_2023} which can adequately reconstruct chest X-rays \cite{chambon_roentgen_2022}. The \ac{VAE} is frozen, and the denoising U-Net is trained on three datasets downsampled and centre-cropped to 512~$\times$~512 pixels: MIMIC-CXR \cite{johnson_mimic-cxr_2019}, ChestX-ray8 \cite{wang_chestx-ray8_2017}, and CheXpert \cite{irvin_chexpert_2019}, totalling \num{487680} training images. This data diversity allows us to perform \emph{zero-shot edits} on datasets not seen during training. 


For MIMIC-CXR, we only include frontal view chest X-rays, and condition the denoising U-Net on the corresponding impression section in the radiology report (a short clinically actionable outline of the main findings). We employ the tokeniser and frozen text encoder from \biovil \cite{bannur_learning_2023}.
For ChestX-ray8 and CheXpert, we condition on a list of all abnormalities present in an image as indicated by the labels, e.g., \prompt{Cardiomegaly. Pneumothorax.}.
If the list of abnormalities is empty, we use the string \prompt{No findings}.
An overview of the labels for each dataset alongside more details on the diffusion model training can be found in \cref{app:details_diff_model}, and more experimental details for the following sections in \crefrange{app:details_covid_exp}{app:details_seg_exp}.

\begin{figure}[t]
    \vspace{-1.1em}
    \centering
    \begin{minipage}{0.64\linewidth}
    \captionof{table}{
        \textbf{Quantifying robustness of \covid detectors to acquisition shift}. We train a weak predictor on the `Biased' dataset---a combination of BIMCV+ \cite{vaya_bimcv_2020} and MIMIC-CXR \citep{johnson_mimic-cxr_2019}; and a strong predictor on an unbiased dataset---a combination of BIMCV+ and \mbox{BIMCV-}; the `Synthetic' test set consists of 2774 \covid-negative images with the same spurious features as the BIMCV+ datasets, e.g. laterality markers. We report mean accuracy and standard deviation across 5 runs. \label{tab:covid_results}
    }
    \end{minipage}
    \hfill
    \begin{minipage}{0.33\linewidth}
    \small
    \begin{tabular}{
        l 
        l 
        S[table-format=2.1(2)] 
    }
        \toprule
        \textbf{Predictor} & \textbf{Test data} & \textbf{Accuracy} \\ 
        \midrule
        Weak   & Biased     & 99.1 \pm 0.2 \\
        Weak   & Synthetic  &  5.5 \pm 2.1 \\
        \addlinespace
        Strong & Biased     & 74.4 \pm 3.0 \\
        Strong & Synthetic  & 76.0 \pm 7.7 \\
        \bottomrule
    \end{tabular}
    \end{minipage}
    \vspace{-2.2em}
\end{figure}

\subsection{Acquisition shift}\vspace{-0.2em}
\label{sec:covid_exp}

\paragraph{Background} In this section, we show how \methodname can be used to quantify the robustness of models to acquisition shift. We closely follow the experimental setup of \citet{degrave_ai_2021}, who show that deep learning systems built to detect \covid from chest X-rays rely on confounding factors rather than pathology features. This problem arises when \covid-positive and -negative images come from disparate sources. In our setup, all \covid-positive cases come from the BIMCV dataset \cite{vaya_bimcv_2020} (denoted BIMCV+), and all \covid-negative cases from MIMIC-CXR \cite{johnson_mimic-cxr_2019} (see \cref{fig:mimic_vs_bimcv}). A model trained on these datasets to classify \covid will rely on spurious features indicative of the data's origin, e.g., laterality markers or the field of view, instead of learning visual features caused by the pathology.

\vspace{-0.6em}

\begin{figure}[b]
    \vspace{-1em}
    \centering
    \captionsetup[subfigure]{justification=centering, skip=-0.7em}
    \begin{minipage}{0.31\linewidth}
        \caption{Removing \covid features with LANCE\footref{foot:lance} (b) also changes the laterality markers and reduces contrast. In contrast, \methodname (c; ours) preserves anatomical structures and laterality markers, and retains the original contrast.\vspace{-1.2em} \label{fig:covid_no_mask_example0}}
    \end{minipage}
    \hfill
    \begin{minipage}{0.665\linewidth}
        \includegraphics[width=\linewidth]{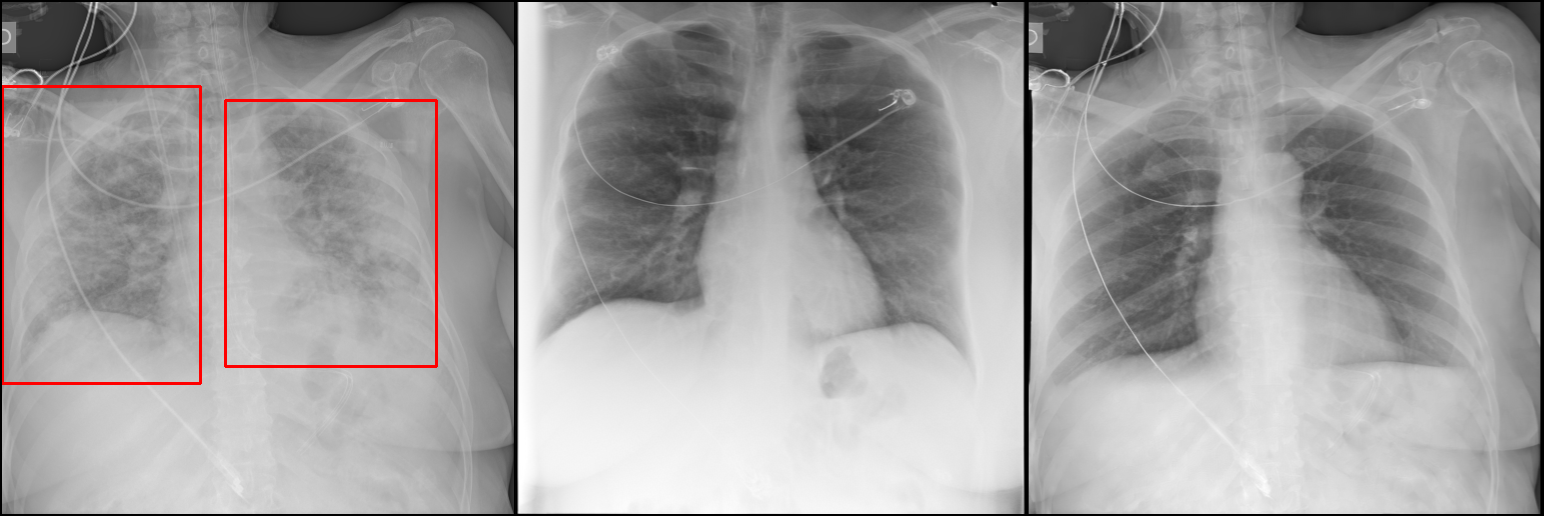}
        \hfill\subcaptionbox{Original Image}[0.32\linewidth]{\centering}
        \hfill\subcaptionbox{LANCE \cite{prabhu_lance_2023}}[0.32\linewidth]{\centering}
        \hfill\subcaptionbox{\methodname (ours)\label{subfig:covid-ours}}[0.32\linewidth]{\centering}
        \hfill\null
    \end{minipage}
    \vspace{-0.3em}
\end{figure}

\paragraph{Setup} A synthetic test set is created by applying \methodname to remove \covid features\footref{foot:terms} from BIMCV+ images using the prompt \prompt{No acute cardiopulmonary process}\footnote{\label{foot:prompt}This is a common radiological description of a `normal' chest X-ray.} (\cref{fig:covid_no_mask_example0}); the included bounding boxes of \covid features are used as the edit mask $m_\text{edit}$. Since this is the only mask available, we set the keep mask as $m_\text{keep}\!=\!1\!-\!m_\text{edit}$. After filtering using the \biovil editing (\cref{sec:clip-filtering}), this results in a synthetic dataset of 2774 \covid-negative images containing the same spurious features as BIMCV+. 

\footnotetext[3]{\label{foot:lance}For LANCE, we perform the text perturbation manually.}

\vspace{-0.6em}

\paragraph{Findings} \cref{tab:covid_results}, shows the performance of a \covid classifier (weak predictor) trained on BIMCV+ and MIMIC-CXR.
In accordance with \citet{degrave_ai_2021}, we find that the weak predictor performs exceptionally well on the real test set (i.e. test splits of both datasets) since the model learned to distinguish the two data sources instead of learning visual features related to \covid.
However, in the second row of \cref{tab:covid_results}, we see a drop of 95\% in accuracy meaning that the model fails to classify the synthetic images as \covid-negative. The weak predictor is not robust to a shift in acquisition.


To show that the decreased performance of the weak predictor is not caused by artefacts in the edited images, we train a more robust \covid classifier (strong predictor), using the BIMCV+ and BIMCV- datasets, as in \cite{degrave_ai_2021}, where the BIMCV- dataset consists of only \covid-negative cases from BIMCV, and test on the same two test datasets. Comparing rows one and three of \cref{tab:covid_results}, we find that the strong predictor performs worse on the test set containing samples from BIMCV+ and MIMIC-CXR than the weak predictor (row one). This is expected as the strong predictor relies on actual pathology features. Lastly, rows three and four of \cref{tab:covid_results} show that the strong predictor performs similarly on the real and synthetic test sets, attesting the quality of our edits.

\vspace{-0.2em}\subsection{Manifestation shift}\label{sec:ptx_exp}\vspace{-0.1em}
\paragraph{Background} In this section, we show how \methodname can be used to quantify the robustness of biomedical vision models to manifestation shift. We closely follow the experimental setup of \citet{rueckel_pneumothorax_2021}, who demonstrate that pneumothorax\footref{foot:terms} classification models are strongly biased by the presence of chest drains: while the average performance of pneumothorax classifiers is high, performance on the subset of images with a chest drain but no pneumothorax is significantly lower. This is due to chest drains being a common treatment for pneumothorax, resulting in the majority of images in datasets like CANDID-PTX \cite{feng_curation_2021} containing a chest drain only if there is a pneumothorax. As a result, only 1\% of images in CANDID-PTX contain a chest drain but no pneumothorax. 

\vspace{-0.5em}

\begin{figure}[t]
    \vspace{-1em}
    \centering
    \begin{minipage}{0.63\linewidth}
    \captionof{table}{
        \textbf{Quantifying robustness of pneumothorax detectors to manifestation shift}. The weak predictor is trained on the biased CANDID-PTX \cite{feng_curation_2021} dataset to classify pneumothorax; the strong predictor is trained on SIIM-ACR \cite{siim-acr-pneumothorax-segmentation} to classify and segment the pneumothorax. Real `Biased' test data comes from CANDID-PTX which exhibits strong confounding between the pneumothorax and chest tubes; `Synthetic' test data is 629 solely edited images containing chest drains but no pneumothorax. We report mean accuracy and standard deviation across 5 runs. \label{tab:ptx_results}}
    \end{minipage}
    \hfill
    \begin{minipage}{0.33\linewidth}
     \small
     \begin{tabular}{
        l 
        l 
        S[table-format=2.1(2)] 
    }
        \toprule
        \textbf{Predictor} & \textbf{Test data} & \textbf{Accuracy} \\ 
         \midrule
        Weak   & Biased    & 93.3 \pm 0.6 \\
        Weak   & Synthetic & 17.9 \pm 3.7 \\
        \addlinespace
        Strong & Biased    & 93.7 \pm 1.3 \\
        Strong & Synthetic & 81.7 \pm 7.1 \\
         \bottomrule
    \end{tabular}
    \end{minipage}
    \vspace{-0.5em}
\end{figure}

\begin{figure}[t]
    \vspace{-0.2em}
    \centering
    \captionsetup[subfigure]{justification=centering, skip=-0.7em}
    \begin{minipage}{0.31\linewidth}
        \caption{Removing pneumothorax (\textcolor{myred}{red}) with LANCE\footref{foot:lance} (b) also removes the spuriously correlated chest drain (\textcolor{myblue}{blue}) and reduces contrast. In contrast, \methodname (c; ours) preserves the chest drain and better preserves anatomical structures.\vspace{-0.8em} \label{fig:no_mask_0}}
    \end{minipage}
    \hfill
    \begin{minipage}{0.66\linewidth}
        \includegraphics[width=\linewidth]{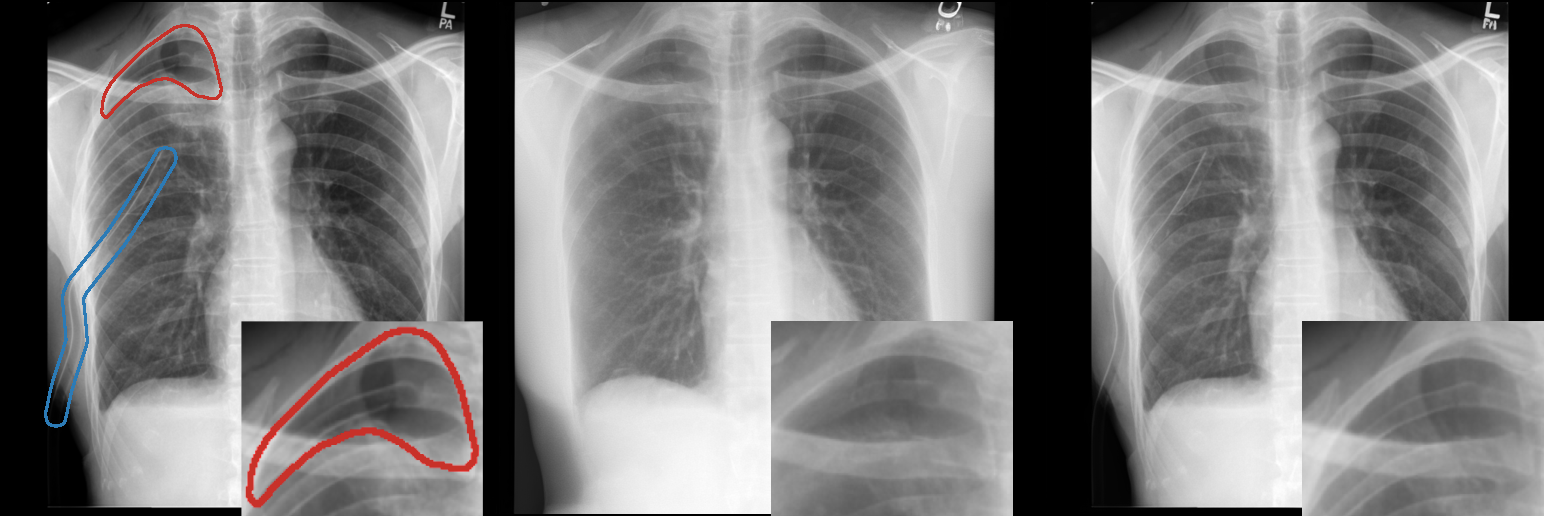}
        \hfill\subcaptionbox{Original Image \label{subfig:pneumothorax-original}}[0.32\linewidth]{\centering}
        \hfill\subcaptionbox{LANCE \cite{prabhu_lance_2023} \label{subfig:pneumothorax-lance}}[0.32\linewidth]{\centering}
        \hfill\subcaptionbox{\methodname (ours) \label{subfig:pneumothorax-ours}}[0.32\linewidth]{\centering}
        \hfill\null
        
    \end{minipage}
    \vspace{-1.4em}
\end{figure}

\paragraph{Setup} We use \methodname to create a synthetic dataset containing images with a chest drain but no pneumothorax, by taking images from CANDID-PTX and editing out the pneumothorax using the prompt \prompt{No acute cardiopulmonary process}\footref{foot:prompt} (\cref{fig:no_mask_0}). The edit mask $m_\text{edit}$ is set as a mask of the pneumothorax, and the keep mask $m_\text{keep}$ is set as the chest drain mask. This ensures that the chest drain will still be present after editing, while allowing the rest of the image to change, preventing border artefacts. After filtering using the \biovil editing score (\cref{sec:clip-filtering}), 628 images are left; in contrast, the real test set contains only 16 of cases with drains but no pneumothorax.


\vspace{-0.5em}

\paragraph{Findings} In accordance with \cite{rueckel_pneumothorax_2021}, we show in \cref{tab:ptx_results} that a pneumothorax classifier (weak predictor) trained on CANDID-PTX performs exceptionally well on the test split of CANDID-PTX, since very few images contain a chest drain and no pneumothorax. However, in row two of \cref{tab:ptx_results}, we show a drastic drop in performance on the synthetic test set, i.e., the weak predictor is not robust to manifestation shift. To show that the drop in performance on the synthetic dataset does not come from editing artefacts, we also train a more robust model (strong predictor) on SIIM-ACR \cite{siim-acr-pneumothorax-segmentation}, following \citet{rueckel_pneumothorax_2021}. The strong predictor is trained to detect the presence of pneumothorax, as well as to segment pneumothorax and chest drains. Testing the strong predictor on the same test datasets (rows three and four of \cref{tab:ptx_results}), we find that the strong predictor performs on par with the weak predictor in row one; however, the strong predictor closes the majority of the gap between the real test set and the synthetic one, attesting the quality of our edits. In agreement with \citet{rueckel_pneumothorax_2021}, there is still a performance gap, indicating that the strong predictor still suffers from mild manifestation shift.

\vspace{-0.3em}
\subsection{Population shift}\vspace{-0.3em}
\label{sec:seg_exp}


\paragraph{Background} In this section, we show how \methodname can be used to quantify the robustness of lung segmentation models to population shifts. Manually segmenting X-ray images is labour intensive and requires high expertise, leading to small datasets often limited to single pathologies or healthy patients \cite{shiraishi_development_2000, jaeger_two_2014}, e.g., MIMIC-Seg \cite{chen_chest_2022}. These models are thus sensitive to occlusions such as medical devices or pathologies, which typically appear as white regions on X-rays \cite{liu_automatic_2022}. Evaluating model robustness requires further image collection for each occlusion type, which is time-consuming and costly.

\vspace{-0.6em}

\begin{figure}[b]
    \vspace*{-0.5em}
    \centering
    \captionsetup[subfigure]{justification=centering, skip=-0.7em}
    \begin{minipage}{0.4\linewidth}
        \caption{Adding pulmonary edema (top), pacemakers (middle), and consolidation (bottom) with \methodname. The `strong predictor' (d), a segmentation model trained on CheXmask \citep{gaggion_chexmask_2023} (a large dataset containing various abnormalities) is more robust to these abnormalities than the `weak predictor' (c), a segmentation model trained on MIMIC-Seg \citep{chen_chest_2022} (a small set of mostly healthy patients): the weak predictor traces around the pacemaker and poorly annotates the consolidated lung. \textcolor{myblue}{Blue}: ground-truth annotation; \textcolor{myred}{red}:~predicted segmentation.}
        \label{fig:segmentation-examples}
    \end{minipage}
    \hfill
    \begin{minipage}{0.56\linewidth}
        \includegraphics[width=\linewidth]{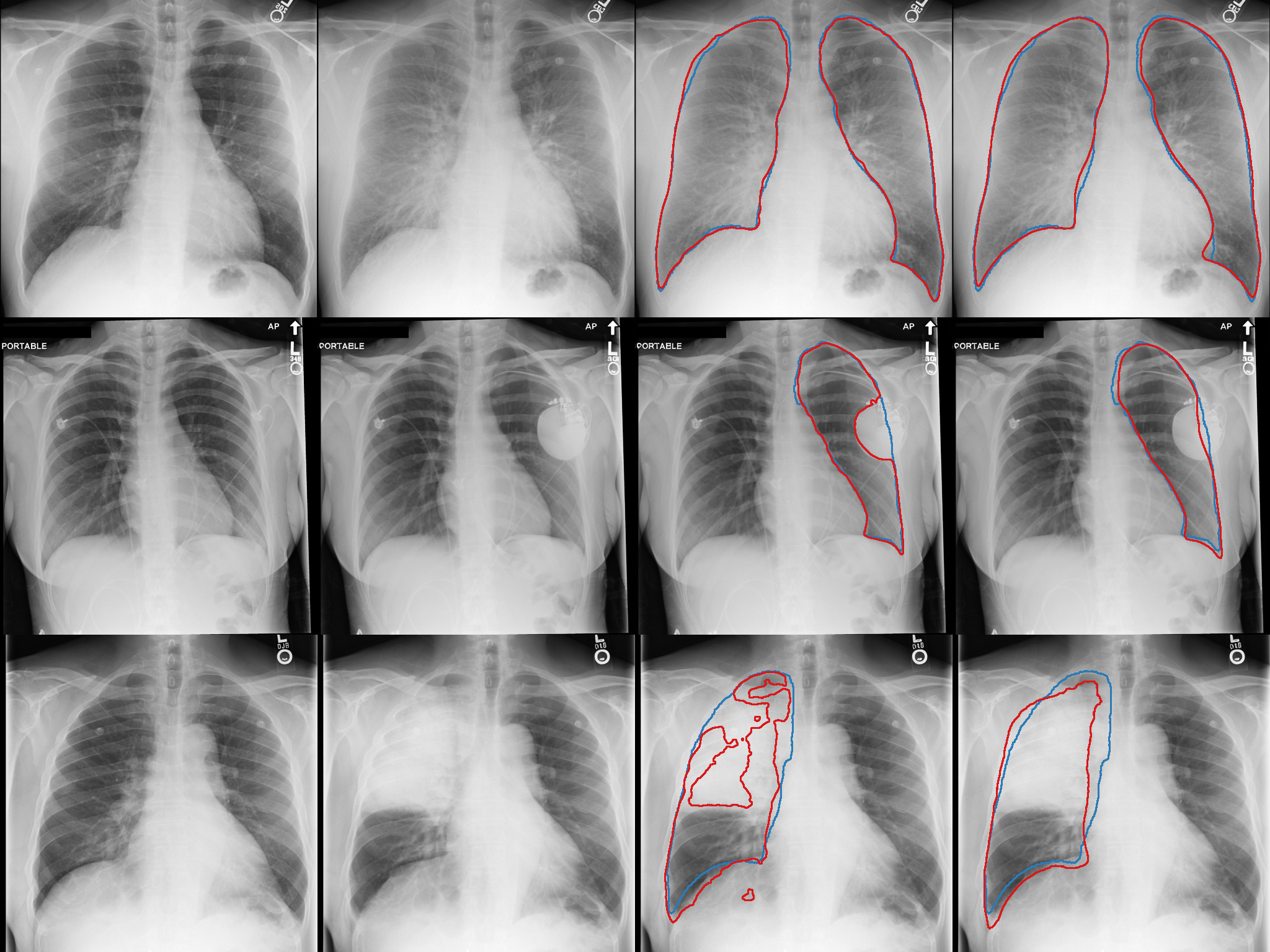}
        \hfill\subcaptionbox{\scriptsize{Original}}[0.24\linewidth]{\centering}
        \hfill\subcaptionbox{\scriptsize{Edited}}[0.24\linewidth]{\centering}
        \hfill\subcaptionbox{\scriptsize{Weak\\predictor}}[0.24\linewidth]{\centering}
        \hfill\subcaptionbox{\scriptsize{Strong\\predictor}}[0.24\linewidth]{\centering}
        \hfill\null
    \end{minipage}
    \vspace{-1.7em}
\end{figure}


\paragraph{Setup} \methodname allows us to stress-test segmentation models while bypassing the need to collect and label more data. Here, abnormalities are added to the lung region in healthy X-rays from MIMIC-Seg (\cref{fig:segmentation-examples}). Editing is constrained to be within the lungs, meaning that the lung boundaries should remain unchanged after editing, by setting the edit masks $m_{\text{edit}}$ as the ground-truth lung segmentations. When editing a single lung, the keep mask $m_{\text{keep}}$ corresponds to the lung which must not change, while when editing both lungs we set $m_{\text{keep}}\!=\!0$. This allows the region outside of the lungs to potentially change to allow opacity adjustments, or for elements to be added outside of the lungs. Stress-test sets are generated for three abnormalities: pulmonary edema, pacemakers, and consolidation\footref{foot:terms}. Prompts are phrased to match similar impressions in the training data (see \cref{app:details_seg_exp}). We evaluate segmentation quality using Dice similarity coefficient, which is the harmonic mean of the precision and recall, and 95th percentile \ac{AHD}, a measure of the distance between two sets \citep{maierhein2023metrics}.

\vspace{-0.5em}




\begin{table}[t] 
    \sisetup{round-precision=1,round-mode=places}
     \caption{
         \textbf{Quantifying robustness of lung segmentation models to population shift}. The `weak predictor' is trained on MIMIC-Seg (a small set of predominantly healthy patients); the `strong predictor' is trained on CheXmask (a large mixed set of patients with various abnormalities). Synthetic test data is created by using \methodname to add edema, pacemakers, and consolidation. We report the change ($\Delta$) in Dice score and \ac{AHD} with respect to the segmentation models evaluated on the ground-truth test set.
     }
    \centering
    \small
    \vspace{-0.5em}
    \begin{tabular}{
        @{}
        l
        S[table-format=2.1]
        S[table-format=2.1]
        @{\hskip 1.4em}
        S[table-format=2.1]
        S[table-format=2.1]
        @{\hskip 1.4em}
        S[table-format=2.1]
        S[table-format=2.1]
        @{\hskip 1.4em}
        S[table-format=2.1]
        S[table-format=2.1]
        @{}
    } 
         \toprule
                  & \multicolumn{4}{c}{\textbf{Weak Predictor}} & \multicolumn{4}{c}{\textbf{Strong Predictor}} \\
        \cmidrule(lr{1.95em}){2-5}\cmidrule(lr){6-9}
        \textbf{Test data} & \textbf{Dice} $\boldsymbol\uparrow$ & $\boldsymbol\Delta$ $\boldsymbol\downarrow$ & \textbf{\acs{AHD}} $\boldsymbol\downarrow$ & $\boldsymbol\Delta$ $\boldsymbol\downarrow$ & \textbf{Dice} $\boldsymbol\uparrow$ & $\boldsymbol\Delta$ $\boldsymbol\downarrow$ & \textbf{\acs{AHD}} $\boldsymbol\downarrow$ & $\boldsymbol\Delta$ $\boldsymbol\downarrow$ \\ 
         \midrule
        Real data & 97.4 & \text{---} & 6.07 & \text{---} & 95.5 & \text{---} & 11.62 & \text{---} \\ 
        Healthy $\overset{\mbox{\tiny edit}}{\rightarrow}$ edema & 93.8 & 3.6 & 21.79 & 15.72 & 93.9 & \bfseries 1.6 & 22.79 & \bfseries 11.17 \\
        Healthy $\overset{\mbox{\tiny edit}}{\rightarrow}$ pacemaker & 85.0 & 12.4 & 49.81 & 43.74 & 87.3 & \bfseries 8.2 & 29.53 & \bfseries 17.91  \\
        Healthy $\overset{\mbox{\tiny edit}}{\rightarrow}$ consolidation &  85.9 & 11.5 & 44.13 & 38.06 & 88.1 & \bfseries 7.4 & 29.41 & \bfseries 17.79 \\
         \bottomrule
    \end{tabular}
    \label{tab:segmentation-results}
\end{table}

\paragraph{Findings} \cref{tab:segmentation-results} shows that a lung segmentation model (weak predictor) trained on MIMIC-Seg performs well on the real biased test data, mostly composed of healthy subjects. However, testing on the synthetic lung abnormality datasets (rows two to four), causes performance to drop substantially, i.e. the weak predictor is not robust to population shift. To show that this drop in performance does not come from editing artefacts, we train a more robust segmentation model (strong predictor) on CheXmask \cite{gaggion_chexmask_2023}, a larger dataset with various lung abnormalities. Testing the strong predictor on the synthetic test sets, we see considerably smaller changes in performance. 
This can be seen in \cref{fig:segmentation-examples}: for pulmonary edema, both models can accurately segment, despite the abnormality; for pacemakers, the weak predictor incorrectly segments around the pacemakers, while the strong predictor more accurately segments the lungs; and for consolidation, both models are less able to segment the lungs accurately, however, the strong predictor gets closer to the ground-truth. See \cref{app:details_seg_exp} for more examples.

\vspace{-0.2em}
\subsection{Quantifying the limitations of existing editing methods}
\begin{figure}[t]
    \sisetup{round-precision=1,round-mode=places}
    \centering
    \captionsetup[subfigure]{justification=centering, skip=-0.8em}
    \includegraphics[width=\linewidth]{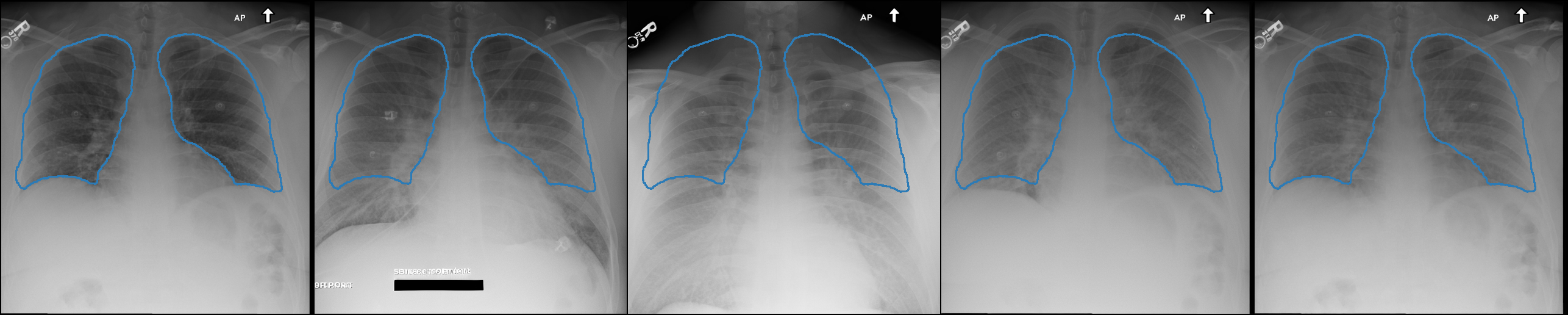}
    \centering
    \hfill\subcaptionbox{}[0.19\linewidth]{\centering}
    \hfill\subcaptionbox{}[0.19\linewidth]{\centering}
    \hfill\subcaptionbox{}[0.19\linewidth]{\centering}
    \hfill\subcaptionbox{}[0.19\linewidth]{\centering}
    \hfill\subcaptionbox{}[0.19\linewidth]{\centering}
    \hfill
    \label{fig:ablation}
    \vspace{-0.8em}
    \noindent\begin{minipage}{0.57\linewidth}
        \vspace{-0.5em}
        \caption{Comparison of LANCE\footref{foot:lance} and \methodname. We measure how well the strong predictor from \cref{tab:segmentation-results}'s outputs matches the ground-truth lung masks (\textcolor{myblue}{blue}) for four synthetic datasets created by adding edema using LANCE and \methodname with DDIM or DDPM inversion. High Dice / low \ac{AHD} indicates that the editing method well preserves the lung border's location and shape. \label{tab:ablation}}
    \end{minipage}
    \hfill
    \begin{minipage}{0.41\linewidth}
        \small
        \begin{tabular}{
                @{}
                l 
                S[table-format=2.1]
                S[table-format=2.1] 
                @{}
            }
                \toprule
                \textbf{Editing Method} & \textbf{Dice} $\boldsymbol\uparrow$  & \textbf{\acs{AHD}} $\boldsymbol\downarrow$  \\ 
                \midrule
                (a) Original data & 95.5 & 11.62 \\
                (b) LANCE w/ DDIM& 78.9  & 65.14 \\
                (c) LANCE w/ DDPM & 80.1  & 69.45 \\
                (d) \methodname w/ DDIM    & 86.2  & 39.83 \\
                (e) \methodname w/ DDPM & \bfseries 93.9 & \bfseries 22.79 \\
                \bottomrule
            \end{tabular}
    \end{minipage}
\end{figure}

\vspace{-0.2em}
\label{sec:quant_ablation}
\paragraph{LANCE}
As seen in the second row of \cref{tab:segmentation-results}, adding edema leads only to a small drop in performance of the strong predictor. We hypothesise that further drops in performances stem from a mismatch of the original mask and the edited images. We therefore use this setup to quantify how well LANCE and \methodname preserve the shape and position of the lung borders. Additionally, we study the difference between results using DDIM or DDPM inversion. For all four methods in \cref{tab:ablation}, we use the same setup as in \cref{sec:seg_exp}: we first edit the original image with the prompt \prompt{Moderate pulmonary edema. The heart size is normal}, and then compare the outputs of the strong predictor with the original ground-truth lung masks. Here, we find that using masks and DDPM inversion is necessary for \methodname to preserve the shape and position of the lung border.

\begin{figure}[b]
    \sisetup{round-precision=1,round-mode=places}
    \vspace{-0.8em}
    \centering
    \begin{subfigure}{0.62\linewidth}
        \includegraphics[width=\linewidth]{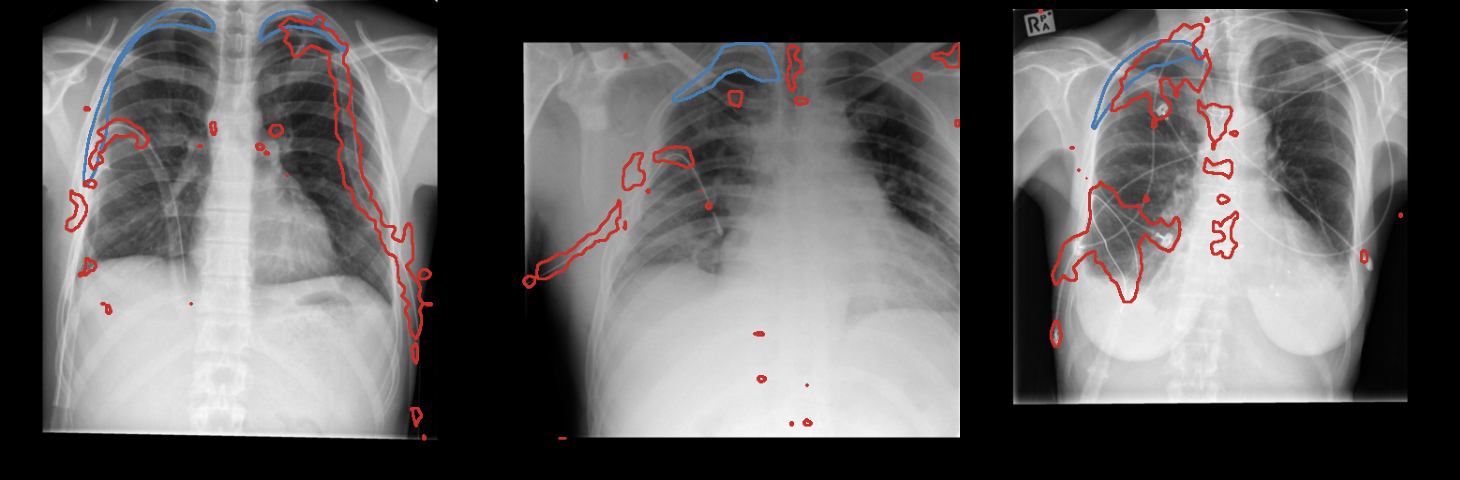}
        \caption{Examples of pneumothorax masks predicted using DiffEdit \cite{couairon_diffedit_2022}. \textcolor{myblue}{Blue}: ground-truth annotation; \textcolor{myred}{red}:~predicted editing mask.\vspace{-0.5em}}
    \end{subfigure}
    \hfill
    \begin{subfigure}{0.36\linewidth}
        \centering
        \small
        \begin{tabular}{
            l 
            S[table-format=2.1]
            S[table-format=3.1] 
        }
            \toprule
            \textbf{Hyperparameters} & \textbf{Dice} & \textbf{\acs{AHD}} \\ 
            \midrule
            Tuned per image    & 33.8  & 97.7 \\
            Tuned on validation & 18.4  &  256.8 \\
            \bottomrule
        \end{tabular}
        \caption{Segmentation metrics for the pneumothorax mask predicted by DiffEdit \cite{couairon_diffedit_2022}, for hyperparameters tuned on the validation set (bottom) and tuned per image (top; which requires ground-truth masks).}
    \end{subfigure}
     \caption{\textbf{Evaluating pneumothorax masks predicted using DiffEdit \cite{couairon_diffedit_2022}}. (a) Predicted masks (\textcolor{myred}{red}) are noisy, with chest drains often incorrectly segmented as well as or instead of the pneumothorax (\textcolor{myblue}{blue}); (b) this is demonstrated quantitatively with low Dice score and high AHD.}
    \label{fig:diffedit-pneumothorax}
    \vspace{-0.2em}
\end{figure}

\vspace{-0.5em}

\paragraph{DiffEdit}
We quantify how well DiffEdit's automatically predicted masks match the manual ground-truth using the same setup as in \cref{sec:ptx_exp}: we take an image containing pneumothorax and a chest drain, and try to remove only the pneumothorax. We create the editing prompt by splitting the original impressions into one part containing a description of the pneumothorax and the other part containing a description of the chest drain, then replace the description of the pneumothorax with \prompt{No pneumothorax}. DiffEdit should therefore predict a mask containing only the pneumothorax. We perform a grid search on the MIMIC-Seg \cite{chen_chest_2022} validation set over DiffEdit's hyperparameters (noise strength and binarising threshold) to optimise pneumothorax segmentation metrics, then evaluate on the training set. In \cref{fig:diffedit-pneumothorax} we see that DiffEdit's predicted masks obtain poor quantitative metrics where parts of the pneumothorax are often missing, and the spuriously correlated chest drain is often included in the predicted mask. As a result, DiffEdit's predicted masks are unsuitable for stress-testing.

\vspace{-0.5em}
\section{Limitations and future work}\vspace{-0.3em}
Despite the encouraging results presented in the paper, \methodname is not without limitations and more work is needed to extend it to more applications. Currently, training datasets and models must be manually analysed to predict potential failure cases, simulate these failures to test the hypothesis, and finally quantitatively evaluate the model; future work could automate such failure mode discovery. Another limitation is that current editing techniques do not enable all types of stress-testing; for example, with current approaches, we are unable to test segmentation models' behaviour to cardiomegaly (enlarged heart) since this would require segmentation maps to be adjusted after editing. However, this could potentially be enabled by enlarging heart segmentations to simulate cardiomegaly and adjusting the ground-truth lung segmentation accordingly.


When using generative editing methods, it is not possible to completely guarantee that unwanted changes will not occur. With \methodname, we minimise this by forcing certain spuriously correlated regions to remain the same, only using \acl{CFG} within the editing mask, and filtering via image--text alignment. Nonetheless, future work improving the editing space to better maintain structure will further help with this issue, but masks will still be necessary to bypass spurious correlations.


When producing simulated stress test sets, several factors affect edit quality. For example, hyperparameters including \acl{CFG} weight, number of inference steps, and time step to encode to. 
Additionally, components of the generative model place restrictions on which edits are possible: the text encoder must well understand specified pathologies to provide informative features to condition the generative model on; similarly, the diffusion model must be able to capture fine details and well cover the data distribution.

Finally, more research is required to develop better approaches for quantifying edit quality for downstream tasks. In particular, observing a change in downstream performance is not necessarily indicative of real-world performance as edit quality may be poor. While the introduced \biovil editing score can be used to quantify edit quality, this introduces reliance on a potentially biased model. Additionally, the \biovil editing score is not suited to detect the artefacts introduced by LANCE and DiffEdit.

\vspace{-0.3em}
\section{Conclusion}\vspace{-0.3em}
In this study, we illustrate the efficacy of generative image editing as a robust tool for stress-testing biomedical vision models. Our focus is on assessing their robustness against three types of dataset shifts commonly encountered in biomedical imaging: acquisition shift, manifestation shift, and population shift. We highlight that one of the significant challenges in biomedical image editing is the correlations learned by the generative model, which can result in artefacts during the editing process. To mitigate these artefacts, \methodname relies on various types of masks to restrict the effects of the editing to certain areas while ensuring the consistency of the edited images. This approach enables us to generate synthetic test sets of high fidelity that exhibit common dataset shifts. We then use these synthetic test sets to identify and quantify the failure modes of biomedical classification and segmentation models. This provides a valuable supplement to explainable AI approaches such as Grad-CAM \cite{selvaraju_grad_2017} and saliency maps \cite{simonyan_deep_2013, adebayo_sanity_2018}.

%
%
\bibliographystyle{plainnat}
\bibliography{references_non_mendeley}

\newpage
\appendix

\begin{center}
    \textbf{\Large Supplementary Material}

    \vspace{1em}

    \textbf{\Large \methodname: stress-testing biomedical vision models via diffusion image editing}
\end{center}

\section{Medical terminology}
\label{app:terms}
With our editing approach being readily applicable to many (non-medical) applications, we tried our best to keep the paper as accessible as possible to a wider audience, using only a small number of medical terms. In the following section we describe the terms used in more detail.

Note, when interpreting a chest X-ray, it is important to remember that the left and right sides are switched. This is because we view the patient from their anatomical laterality point of view, as if we are facing them. So, what appears on the left in an image is actually the patient’s right side, and vice versa.

\subsection{Pathologies}
\paragraph{Cardiomegaly} This term refers to an enlarged heart, which is usually indicative of an underlying heart condition. The enlargement can include the entire heart, one side of the heart, or a specific area. On a chest X-ray, the heart may appear larger than normal.

\paragraph{Opacities} In the context of a chest X-ray, opacity is a nonspecific descriptor for areas that appear whiter than normal lung. Normally, lungs look dark gray on an X-ray due to presence of air (note the black pure air surrounding the patient on x-ray for reference). If there are whiter areas, it means something is filling up that space inside the lungs, replacing the air.

\paragraph{Pulmonary Edema} is caused by accumulation of fluid in the lungs. In the context of chest X-rays, pulmonary edema appears as increased opacity within and around the air space. In \cref{fig:additional-edema-edits}, we show a variety of pulmonary edema examples.

\paragraph{Consolidation} In the context of chest X-rays, consolidation refers to a region of the lung where the air spaces are filled with fluid, cells, tissue, or other substances. This results in a white region on the X-ray. In \cref{fig:additional-consolidation-edits}, we show a variety of consolidation examples. 

\paragraph{\covid} refers to pneumonia caused by SARS-CoV-2 virus which manifests most commonly as multifocal, bilateral opacities with predominance in the lower half of the lung.

\paragraph{Pneumothorax} This condition occurs when air leaks into the pleural space (between the lung and chest wall), causing the lung to collapse. It can be a complete lung collapse or a collapse of only a portion of the lung. On a chest X-ray, a pneumothorax is seen as a dark region around the edge of the lung, lacking any white texture (except the ribs). The border of the collapsed lung can be seen as in \cref{subfig:app-pneumothorax-original} at the inferior contour of the mask. Often small pneumothorax can be hard to spot on a chest X-ray which contributed to computer vision models overly relying on chest drains for detection, see \cref{sec:ptx_exp}.

\subsection{Support devices}
\paragraph{Chest drain} This is a tube inserted into the pleural space to remove unwanted air (pneumothorax) or fluid (pleural effusion). On an X-ray, you can see the tube in the form of two parallel thin white lines. Its position depends on what it is treating: for pneumothorax it is aimed towards the top; if it is draining fluid, it is towards the bottom.

\paragraph{Pacemaker} This is a device placed under the skin near the collarbone. It helps control abnormal heart rhythms. It has two parts: a control unit (battery and electronics) and wires (white lines) that connect to the heart. In \cref{fig:additional-pacemaker-edits}, we show a variety of pacemaker examples. 

\section{Details for the limitations of LANCE}
\label{app:artefacts_lance}
\begin{figure}[h]
        \centering
        \captionsetup[subfigure]{justification=centering, skip=-0.7em}
        \includegraphics[width=\linewidth]{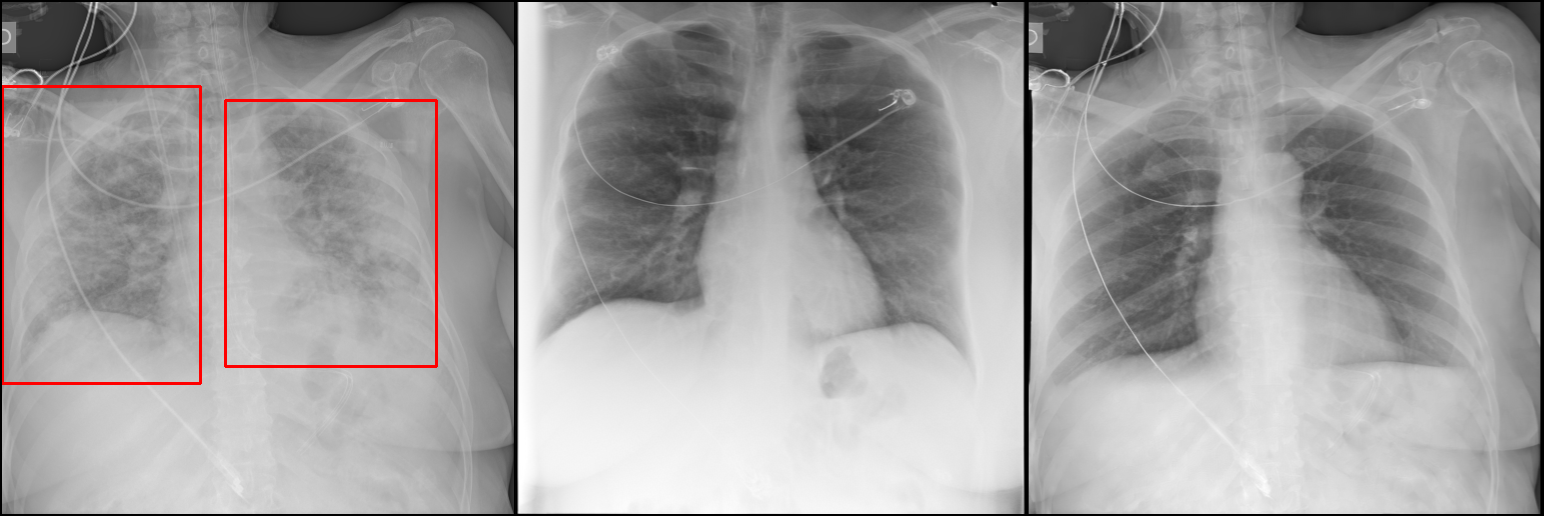}
        \hfill\subcaptionbox{Original Image}[0.32\linewidth]{\centering}
        \hfill\subcaptionbox{LANCE \cite{prabhu_lance_2023}}[0.32\linewidth]{\centering}
        \hfill\subcaptionbox{\methodname (ours)}[0.32\linewidth]{\centering}
        \hfill\null
    \caption{Using LANCE (b) to remove \covid features (rectangle in (a)),the laterality markers are missing. In addition, the field of view is changed.
    In contrast, \methodname (c; ours) uses masks to preserve laterality markers, which also preserves anatomical structures in the process, and retains the original contrast.}
    \label{fig:covid_no_mask_example1}
\end{figure}

During the development of \methodname, we observed numerous artefacts when editing images from the BIMCV+ or CANDID-PTX datasets without using masks. In both instances, the pathology and the lateral markings or chest tubes were removed, leading to potential misinterpretations of the results if these edited images were used for stress-testing. Note, that instead of using a captioner and perturber as seen in the original implementation of LANCE, we manually select the prompts used for editing.
In \cref{fig:covid_no_mask_example1}, we compare \methodname with LANCE (which does not use masks) in editing images from the BIMCV+ dataset. This comparison follows the same experimental setup as in \cref{sec:covid_exp}. \methodname retains the laterality marker on the left of the image, whereas LANCE completely removes it. In both scenarios, we employ the prompt \prompt{No acute cardiopulmonary process}\footref{foot:prompt} to edit the image.

\begin{figure}
        \centering
        \captionsetup[subfigure]{justification=centering, skip=-0.7em}
        \includegraphics[width=\linewidth]{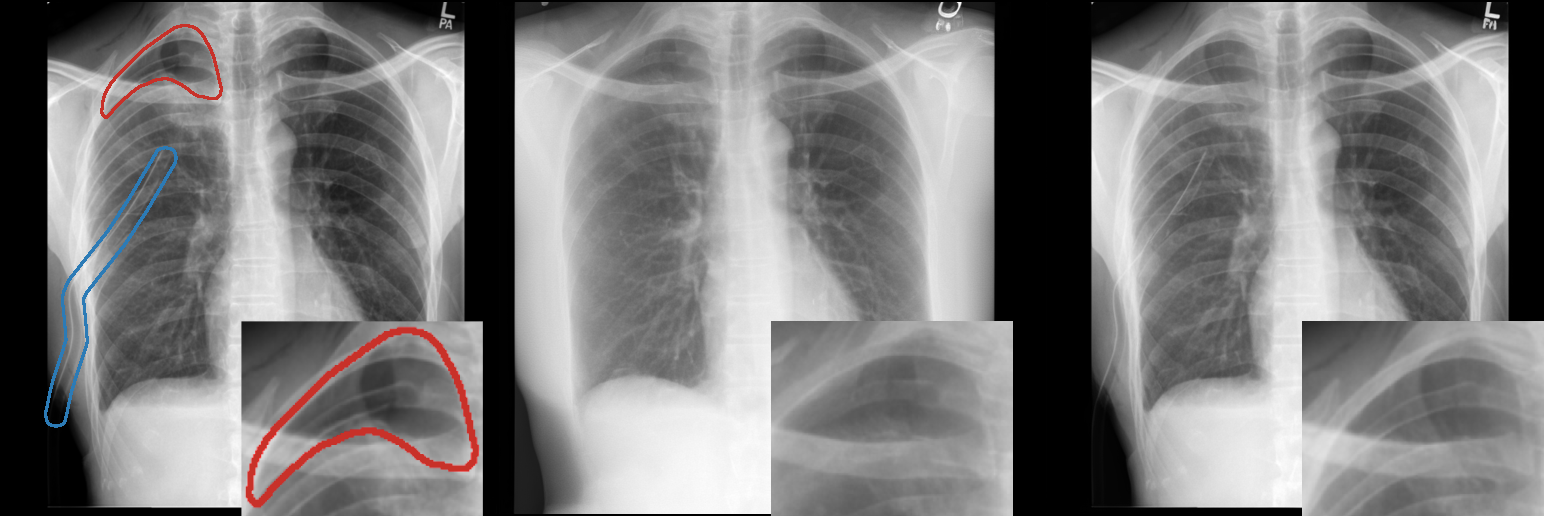}
        \hfill\subcaptionbox{Original Image \label{subfig:app-pneumothorax-original}}[0.32\linewidth]{\centering}
        \hfill\subcaptionbox{LANCE \cite{prabhu_lance_2023} \label{subfig:app-pneumothorax-lance}}[0.32\linewidth]{\centering}
        \hfill\subcaptionbox{\methodname (ours) \label{subfig:app-pneumothorax-ours}}[0.32\linewidth]{\centering}
        \hfill\null
        \vspace*{-0.5em}
        \caption{Removing pneumothorax (\textcolor{myred}{red}) from X-rays using LANCE (b) results in the spuriously correlated chest drain (\textcolor{myblue}{blue}) also being removed. \methodname (c, ours) uses pneumothorax and chest drain masks to remove the pneumothorax while preserving the chest drain. LANCE results in decreased contrast and poorly defined anatomical structures, preserved by \methodname.}
        \label{fig:no_mask}
\end{figure}

Similarly, in \cref{fig:no_mask}, we attempt to remove only the pneumothorax from an image containing a pneumothorax and chest drain, using the prompt \prompt{No acute cardiopulmonary process}\footref{foot:prompt}, while preserving the rest of the image, including the chest drain. For a more comprehensive description of the experimental setup, refer to \cref{sec:ptx_exp}. For LANCE (\cref{subfig:app-pneumothorax-lance}), we note that not only is the region containing the pneumothorax altered, but the chest drain is also removed. This makes LANCE unsuitable for evaluations such as our manifestation shift evaluation (\cref{sec:ptx_exp}), which requires the preservation of support devices like chest drains. We argue that this artefact suggests that the diffusion model has learned correlations between pathologies and support devices, leading to the removal of support devices when prompted to remove a pathology.

\begin{figure}[h]
    \centering
    \captionsetup[subfigure]{justification=centering, skip=-0.7em}
        \includegraphics[width=\linewidth]{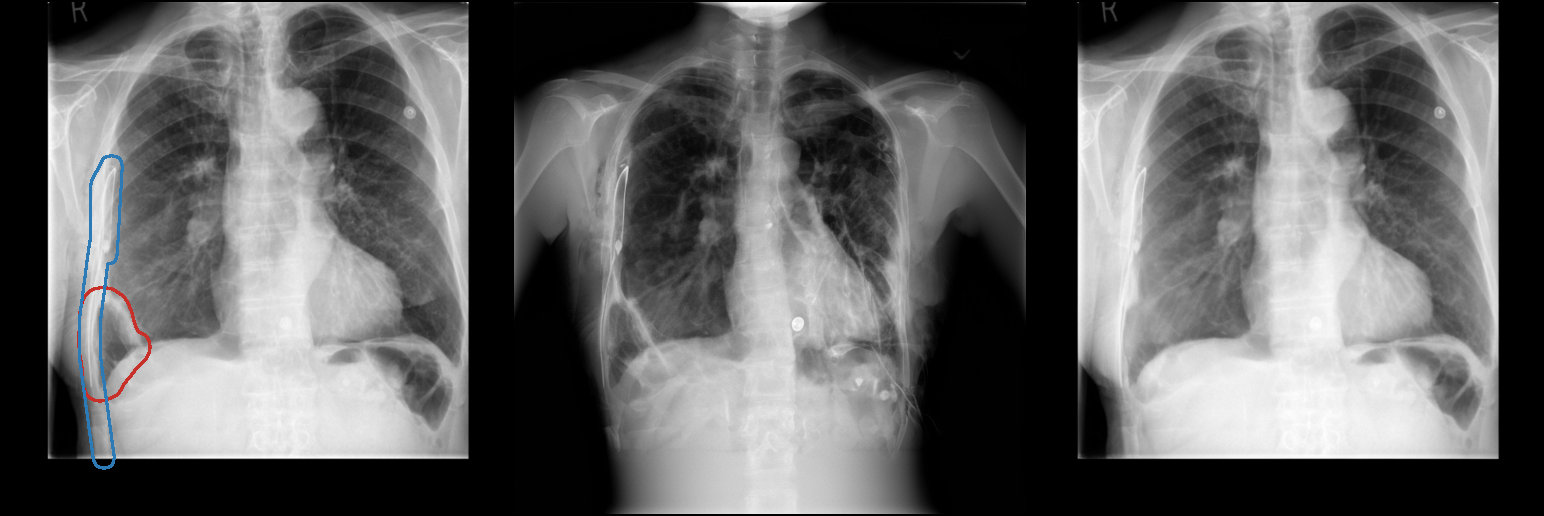}
        \hfill\subcaptionbox{Original Image}[0.32\linewidth]{\centering}
        \hfill\subcaptionbox{LANCE \cite{prabhu_lance_2023}}[0.32\linewidth]{\centering}
        \hfill\subcaptionbox{\methodname (ours)}[0.32\linewidth]{\centering}
        \hfill\null
    \caption{Removing pneumothorax from X-rays using \methodname (c; ours) results in a minimally modified X-ray, with the pneumothorax successfully removed and chest drain still present. In contrast, LANCE (b) fails to properly remove the pneumothorax while keeping most of the chest drain in place, instead modifying the appearance of the drain to look more like a wire; moreover, there are extensive artefacts bilaterally, with abdomen, face, and arms added, modified gas pattern and heart, as well as the lung apexes no longer being asymmetrical, making it unclear whether the X-rays are of the same patient. \textcolor{myblue}{Blue}: ground-truth annotation for chest drain; \textcolor{myred}{red}:~ground-truth annotation for pneumothorax.}
    \label{fig:candid_no_mask_example1}
    \vspace{-0.5em}
\end{figure}

In \cref{fig:candid_no_mask_example1}, we compare \methodname with LANCE in editing images from the CANDID-PTX dataset using the prompt \prompt{No pneumothorax}. We observe that LANCE generates a variety of artefacts. While it retains most of the chest drain, LANCE fails to effectively remove the pneumothorax, instead altering its appearance to resemble a wire. Additionally, there are extensive bilateral artefacts, with modifications to the abdomen, face, and arms, altered gas pattern and heart, and the lung apices no longer being asymmetrical, raising questions about whether the X-rays are from the same patient.

One potential explanation for the artefacts seen in this section is found in recent literature on diffusion models for image-to-image translation. In \citet{su_dual_2023}, the authors show that image-to-image translation can be performed with two independently trained diffusion models. They first obtain a latent representation $\hat{x}_t$ from a source image $x_0$ with the
source diffusion model, and then decode the latent using the target model to construct the target image. We argue that since the diffusion model in \cref{sec:foundation_diff_model} was not trained on data from BIMCV+ or CANDID-PTX, in those cases we perform image-to-image translation along with the image editing. I.e., editing images outside of the training distribution of the diffusion model leads to images that look more similar to images from within the training distribution. In the case of \methodname, where we heavily rely on masks to control the editing, we only observe minor artefacts. However, in the case of LANCE, we observe major artefacts that make LANCE unsuitable for stress-testing of biomedical imaging models. To avoid artefacts, we tried different values for the LANCE hyperparameters, such as the guidance scale, without success.

\section{Details for the limitations of DiffEdit}\label{app:artefacts_diffedit}
In contrast to LANCE, DiffEdit employs a single mask $m_\text{edit}$ for editing. As the editing is only applied within $m_\text{edit}$, DiffEdit avoids the artefacts described in the previous section. However, DiffEdit introduces new artefacts.

In general, DiffEdit consists of two steps. First, it predicts the edit mask $m_\text{edit}$ using the difference between the original prompt and the editing prompt. Second, the editing, following the editing prompt, is applied inside the predicted mask $m_\text{edit}$, leaving the area outside of the mask unchanged. When applying DiffEdit to the experimental setups of \cref{sec:ptx_exp} and \cref{sec:seg_exp} we find problems with both instances.

Initially, we quantify how well the mask automatically predicted by DiffEdit aligns with the ground-truth annotation. We use the same setup as in \cref{sec:ptx_exp}: we take an image containing a pneumothorax and a chest drain (sourced from the CANDID-PTX dataset) and aim to remove only the pneumothorax. We create the editing prompt by splitting the original impressions into one part containing a description of the pneumothorax and another part containing a description of the chest drain. We then replace the part containing the description of the pneumothorax with \prompt{No pneumothorax}. Therefore, DiffEdit should predict a mask containing only the pneumothorax. We perform a grid search on the validation CANDID-PTX dataset over DiffEdit’s hyperparameters, optimising for pneumothorax segmentation metrics, and then evaluate on the training set. In \cref{fig:diffedit-pneumothorax}, we show that masks predicted by DiffEdit obtain poor quantitative metrics compared to the manually annotated masks, where parts of the pneumothorax are often missing, and the spuriously correlated chest drain is often included in the automatically predicted mask. As a result, masks predicted by DiffEdit are unsuitable for editing images that can be used for stress-testing.

\begin{figure}
\includegraphics[width=\linewidth]{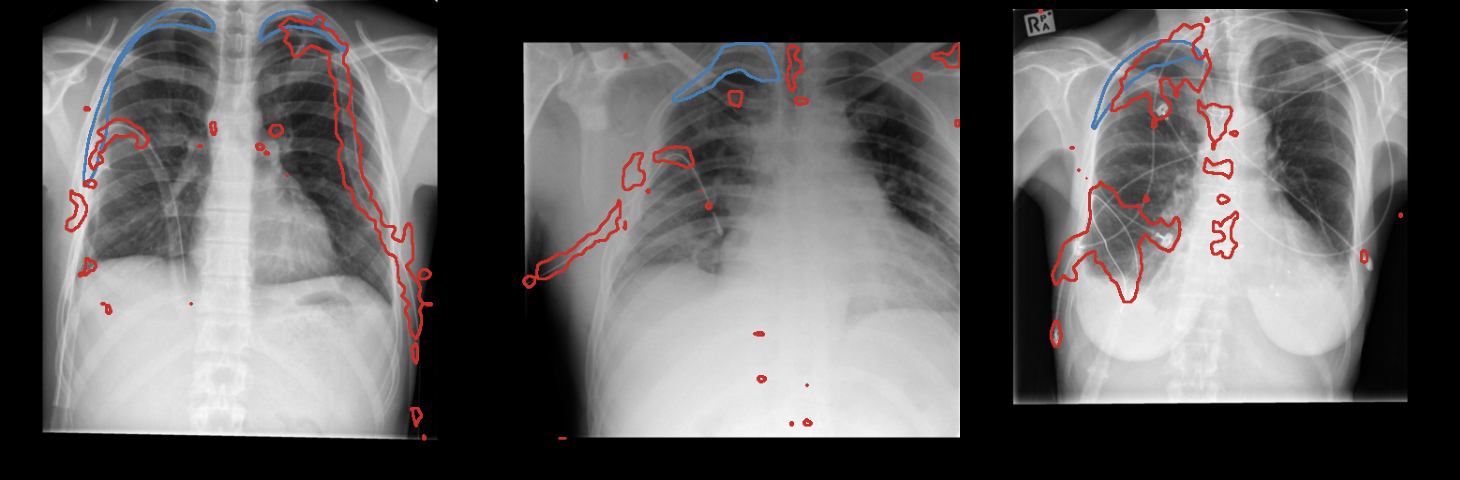}
        \caption{Examples of pneumothorax masks predicted using DiffEdit \cite{couairon_diffedit_2022}. \textcolor{myblue}{Blue}: ground-truth annotation; \textcolor{myred}{red}:~predicted editing mask.}
        \label{fig:diffedit_pred}
\end{figure}

Secondly, in contrast to \methodname, which allows the area outside of the mask to change for consistency, DiffEdit restricts the changes to happen inside the mask. While this would generate valid edits for the experiment in \cref{sec:covid_exp}, it can lead to artefacts in the case of the experiments in \cref{sec:ptx_exp} and \cref{sec:seg_exp}.

\begin{figure}
            \includegraphics[width=\linewidth]{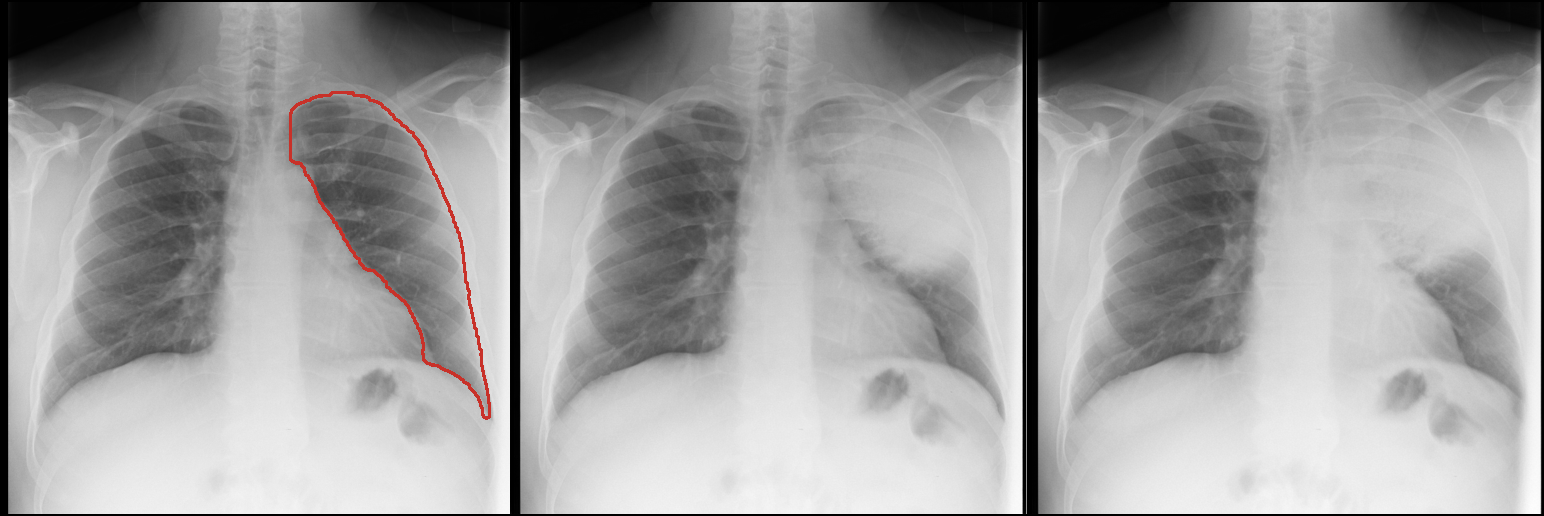}
        \hfill\subcaptionbox{Original Image \label{subfig:consolidation-original}}[0.32\linewidth]{\centering}
        \hfill\subcaptionbox{DiffEdit \cite{couairon_diffedit_2022} \label{subfig:consolidation-diffedit}}[0.32\linewidth]
        \hfill\subcaptionbox{\methodname (ours) \label{subfig:consolidation-ours}}[0.32\linewidth]{\centering}
        {\centering}
        \hfill\null
        \vspace*{-0.5em}
        \caption{Adding consolidation to the left lung using DiffEdit (b) results in a dark border along the original lung mask (\textcolor{myred}{red}) since editing can only occur within the masked region. \methodname (c; ours) allows the region outside of the mask to change to ensure consistency, resulting in more realistic edits. For both editing methods, we use ground-truth masks of the lung.}
        \label{fig:compare_to_diffedit}
    \vspace*{-0.8em}
\end{figure}

Following the setup from \cref{sec:seg_exp}, our goal is to add consolidation to the left upper lung of a healthy patient. In \cref{fig:compare_to_diffedit}, we compare the editing results of \methodname and DiffEdit. While \methodname leads to a realistic occlusion of the heart, DiffEdit fails to generate a realistic-looking edit. Instead, it creates a visible gap between the consolidation and the heart border, which makes the edited image unsuitable for stress-testing a lung segmentation model.

\section[Experimental details for Section 5.1: diffusion model]{Experimental details for \Cref{sec:foundation_diff_model}: diffusion model}
\label{app:details_diff_model}
In this section, we provide additional details on how the diffusion model used for all experiments in \cref{sec:experiments} was trained. The \ac{VAE} downsamples the input images by a factor of eight, meaning that the latent space has spatial dimensions 64 $\times$ 64. For the diffusion model, we use the linear $beta$ schedule and $\epsilon$-prediction proposed by \citet{ho_denoising_2020}. The U-Net architecture is as used by \citet{rombach_high-resolution_2022}, which we instantiate with base channels 128, channel multipliers (1, 2, 4, 6, 8), and self-attention at feature resolutions 32 $\times$ 32 and below, with each attention head being 32-dimensions. The \biovil text encoder \citep{bannur_learning_2023} has a maximum token length of 128, so sentences within the impression are shuffled and then clipped to this length. An exponential moving average is used on model parameters, with a decay factor of 0.999. We drop the text conditioning with $p = 0.1$ during training to allow \ac{CFG} when sampling \cite{ho_classifier-free_2022}. Training was performed using 48 V100 GPUs for 300 epochs using automatic mixed precision. The AdamW \citep{loshchilov_coupled_2018} optimiser was used, with a fixed learning rate of 10$^{-4}$.

The preprocessing steps are:
\begin{enumerate}[itemsep=1pt, parsep=1pt, topsep=1pt]
    \item Resize such that the short side of the image has size 512, using bilinear interpolation;
    \item Centre-crop to 512 $\times$ 512 pixels;
    \item Map minimum and maximum intensity values to $[-1, 1]$.
\end{enumerate}

We use the following label categories for the CheXpert dataset:\vspace{-0.5em}
\begin{enumerate}
    \begin{minipage}[t]{0.49\linewidth}
        \item Atelectasis
        \item Cardiomegaly
        \item Consolidation
        \item Edema
        \item Enlarged\\cardiomediastinum
        \item Fracture
        \item Lung lesion
    \end{minipage}
    \begin{minipage}[t]{0.49\linewidth}
        \item Lung opacity
        \item No finding
        \item Pleural effusion
        \item Pleural other
        \item Pneumonia
        \item Pneumothorax
        \item Support devices
    \end{minipage}
\end{enumerate}

For ChestX-ray8, we use:\vspace{-0.5em}
\begin{enumerate}
    \begin{minipage}[t]{0.49\linewidth}
        \item Atelectasis
        \item Cardiomegaly
        \item Consolidation
        \item Edema
        \item Effusion
        \item Emphysema
        \item Fibrosis
        \item Hernia
    \end{minipage}
    \begin{minipage}[t]{0.49\linewidth}
        \item Infiltration
        \item Mass
        \item No Finding
        \item Nodule
        \item Pleural thickening
        \item Pneumonia
        \item Pneumothorax
    \end{minipage}
\end{enumerate}

\section[Experimental details for Section 5.2: acquisition shift]{Experimental details for \Cref{sec:covid_exp}: acquisition shift}
\label{app:details_covid_exp}

The datasets used and their respective train / validation / test splits are as follows:

\begin{enumerate}[itemsep=1pt, parsep=1pt, topsep=1pt]
    \item BIMCV+: 3008 / 344 / 384
    \item BIMCV-: 1721 / 193 / never used for testing
    \item MIMIC-CXR: 5000 / 500 / 500 (randomly sampled)
    \item Synthetic: never used for training or validation / 2774 (after filtering)
\end{enumerate}
All splits were made ensuring non-overlapping subject IDs.

The filtering of the synthetic test dataset was done using the prompts: \prompt{Opacities} and \prompt{No acute cardiopulmonary process}\footref{foot:prompt}.

For training, we converted the original labels of the BIMCV datasets as follows:
if an image has the label `Negative for Pneumonia' or `Atypical Appearance' we assign label 0; while if it has the label `Typical Appearance' or `Indeterminate Appearance' we assign label 1. 

The classifier is trained using a ResNet50 architecture with batch size 32, 100 epochs and learning rate 10$^{-5}$.
The model was evaluated at the point of best validation \ac{AUROC}.

The preprocessing steps are as in \cref{app:details_diff_model}, but image intensities are mapped to $[0, 1]$.

The following augmentations were used:
\begin{enumerate}[itemsep=1pt, parsep=1pt, topsep=1pt]
    \item Random horizontal flip with probability 0.5
    \item Random affine transformations with rotation $\theta \sim \mathcal{U}(-30, 30)$ degrees and shear $\phi \sim \mathcal{U}(-15, 15)$ degrees
    \item Random colour jittering with brightness $j_b \sim \mathcal{U}(0.8, 1.2)$ and contrast $j_c \sim \mathcal{U}(0.8, 1.2)$
    \item Random cropping with scale $s \sim \mathcal{U}(0.8, 1)$
    \item Addition of Gaussian noise with mean $\mu = 0$ and standard deviation $\sigma = 0.05$
\end{enumerate}

\section[Experimental details for Section 5.3: manifestation shift]{Experimental details for \Cref{sec:ptx_exp}: manifestation shift}
\label{app:details_ptx_exp}
The datasets used and their respective train / validation / test splits are as follows:

\begin{enumerate}[itemsep=1pt, parsep=1pt, topsep=1pt]
    \item CANDID-PTX: \num{13836} / \num{1539} / \num{1865}
    \item SIIM-ACR: \num{10712} / \num{1625} / never used for testing
    \item Synthetic: never used for training or validation / 629 (after filtering)
\end{enumerate}
All splits were made ensuring non-overlapping subject IDs.

The filtering of the synthetic test dataset was done using the prompts: \prompt{Pneumothorax} and \prompt{No acute cardiopulmonary process}\footref{foot:prompt}.

After observing that the contours of the pneumothorax and chest drain masks often do not include the borders of the pneumothorax or chest drain we apply isotropic dilation with a radius of 5. Examples of such dilated masks can be seen in \cref{fig:candid_no_mask_example1} (a).  

For the `Biased' classifier the same model architecture, training hyperparameters and data augmentation are as described in \cref{app:details_covid_exp}

In the case of the `Unbiased' model, a segmentation model is trained using the EfficientNet
U-Net \cite{tan_efficientnet_2019} architecture. We add a single classification layer to the lowest resolution of the U-Net. The segmentation model is trained to segment pneumothorax, and the classifier is used to detect the presence of pneumothorax.

The combined model is trained for 100 epochs with batch size 16, learning rate \SI{5e-4}, and a cosine scheduler with warm-up during the first 6\% of steps.
The model was evaluated at the point of best validation \ac{AUROC} for the pneumothorax classifier.

Data preprocessing and augmentation were as described in \cref{app:details_covid_exp}, with $s \sim \mathcal{U}(0.9, 1.1)$. Additionally, a random elastic transform with scale 0.15 (as implemented in Albumentations \cite{Buslaev_2020}) was used.

\begin{figure}[t]
  \hfill
  \begin{minipage}{0.33\linewidth}
    \centering
    \includegraphics[width=\textwidth]{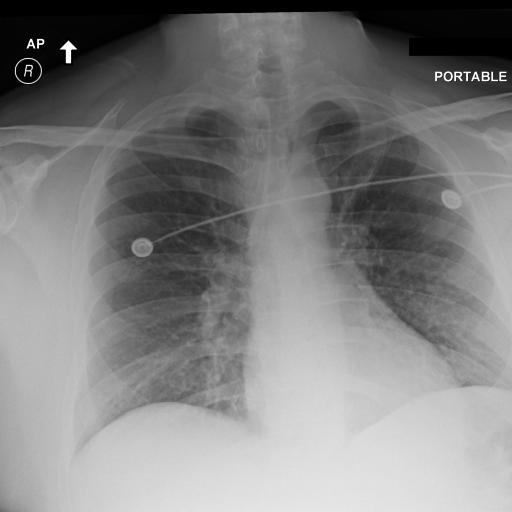}
    \subcaption{Example image from MIMIC-CXR \cite{johnson_mimic-cxr_2019}.}
  \end{minipage}
  \hfill
  \begin{minipage}{0.33\linewidth}  
    \centering
    \includegraphics[width=\textwidth]{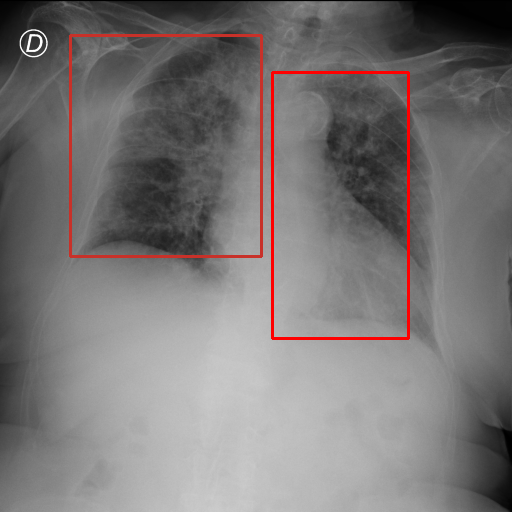}
    \subcaption{Example image from BIMCV+ \cite{vaya_bimcv_2020}.}
  \end{minipage}
  \hfill\null
  \caption{Comparison of the visual appearance between the MIMIC-CXR and BIMCV+ datasets. As shown by \cite{degrave_ai_2021} there are distinct differences in the laterality markings (top left corner) and field of views of the images. Bounding boxes in (b) indicate the presence of abnormalities caused by \covid.}
  \label{fig:mimic_vs_bimcv}
\end{figure}

\section[Experimental details for Section 5.4: population shift]{Experimental details for \Cref{sec:seg_exp}: population shift}
\label{app:details_seg_exp}

Prompts used are as follows:
\begin{itemize}
    \item Pulmonary edema: \prompt{Moderate pulmonary edema. The heart size is normal}
    \item Pacemaker: \prompt{Left pectoral pacemaker in place. The position of the leads is as expected. Otherwise unremarkable chest radiographic examination}
    \item Consolidation: \prompt{New [left/right] upper lobe consolidation}
\end{itemize}

\noindent The datasets used and their respective train / validation / test splits are as follows:

\begin{enumerate}
    \item MIMIC-Seg: 911 / 114 / 115
    \item CheXmask: \num{169206} / \num{36580} / \num{36407}
    \item Synthetic edema: never used for training or validation / 787 (after filtering)
    \item Synthetic Pacemaker: never used for training or validation / 744 (after filtering)
    \item Synthetic Consolidation: never used for training or validation / 1577 (after filtering)
\end{enumerate}
All splits were made ensuring non-overlapping subject IDs.

The same segmentation model architecture, training hyperparameters, and data augmentation/preprocessing steps are used as described in \cref{app:details_ptx_exp}.

\begin{figure}[p]
    \centering
    \captionsetup[subfigure]{justification=centering, skip=-0.7em}
    \begin{minipage}{0.8\linewidth}
        \centering
        \includegraphics[width=\linewidth]{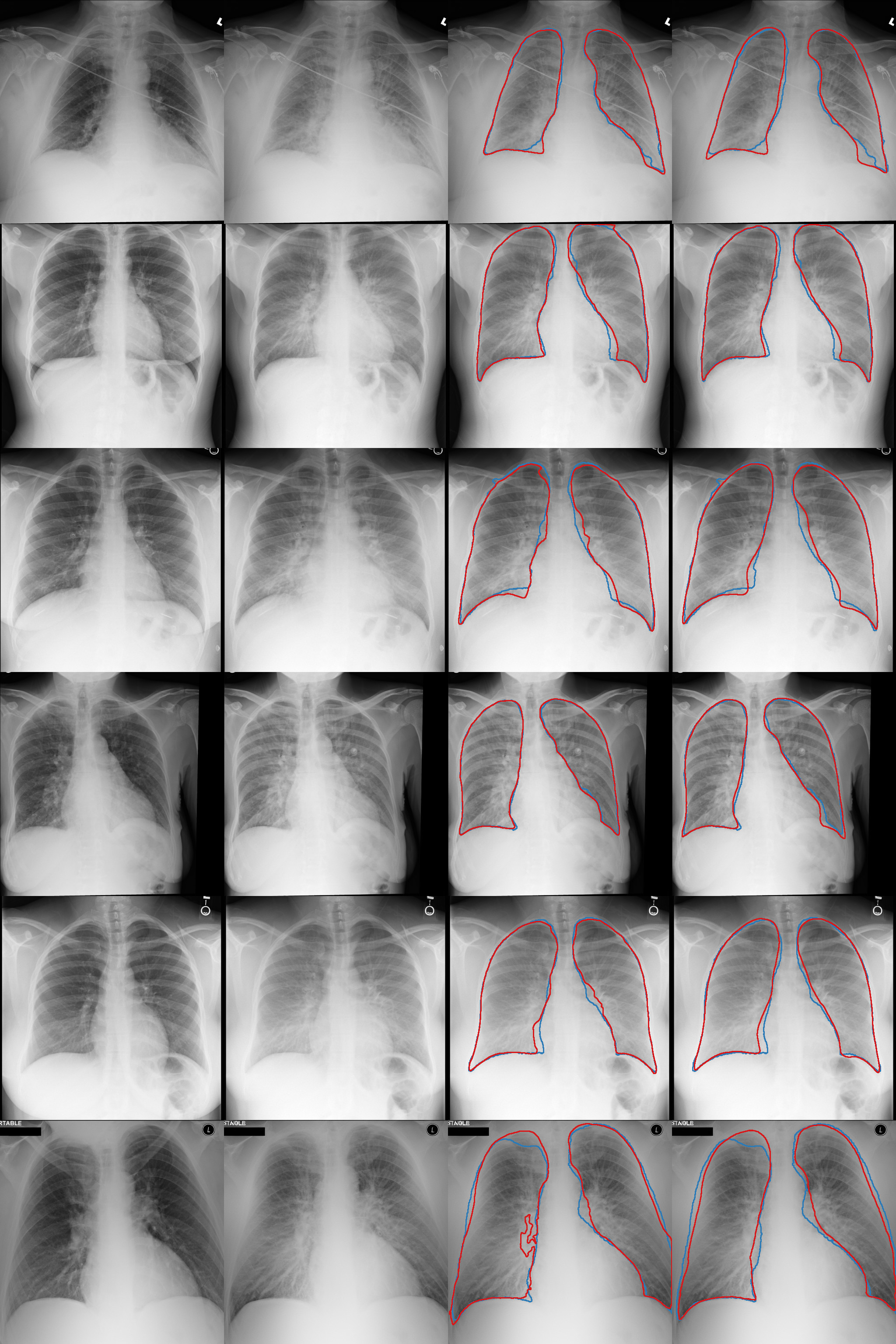}
        \hfill\subcaptionbox{\small{Original}}[0.24\linewidth]{\centering}
        \hfill\subcaptionbox{\small{Edited}}[0.24\linewidth]{\centering}
        \hfill\subcaptionbox{\small{Weak\\Predictor}}[0.24\linewidth]{\centering}
        \hfill\subcaptionbox{\small{Strong\\Predictor}}[0.24\linewidth]{\centering}
        \hfill\null
    \end{minipage}
    \caption{Additional edits simulated by \methodname for stress-testing two segmentation models. The `weak predictor' (c) and the `strong predictor' (d) are trained on MIMIC-Seg \cite{chen_chest_2022} and CheXmask \cite{gaggion_chexmask_2023} respectively, by adding pulmonary edema, via the prompt \prompt{Moderate pulmonary edema. The heart size is normal.} \textcolor{myblue}{Blue}: ground-truth mask: ; \textcolor{myred}{red}: predicted. Similar to the example in \cref{fig:segmentation-examples}, both segmentation models predict relatively accurate segmentation maps, indicating a high level of robustness to this pathology. Edits are visually high quality, with anatomy well maintained, and the edema clearly identifiable.} 
    \label{fig:additional-edema-edits}
\end{figure}

\begin{figure}[p]
    \centering
    \captionsetup[subfigure]{justification=centering, skip=-0.7em}
    \begin{minipage}{0.8\linewidth}
        \includegraphics[width=\linewidth]{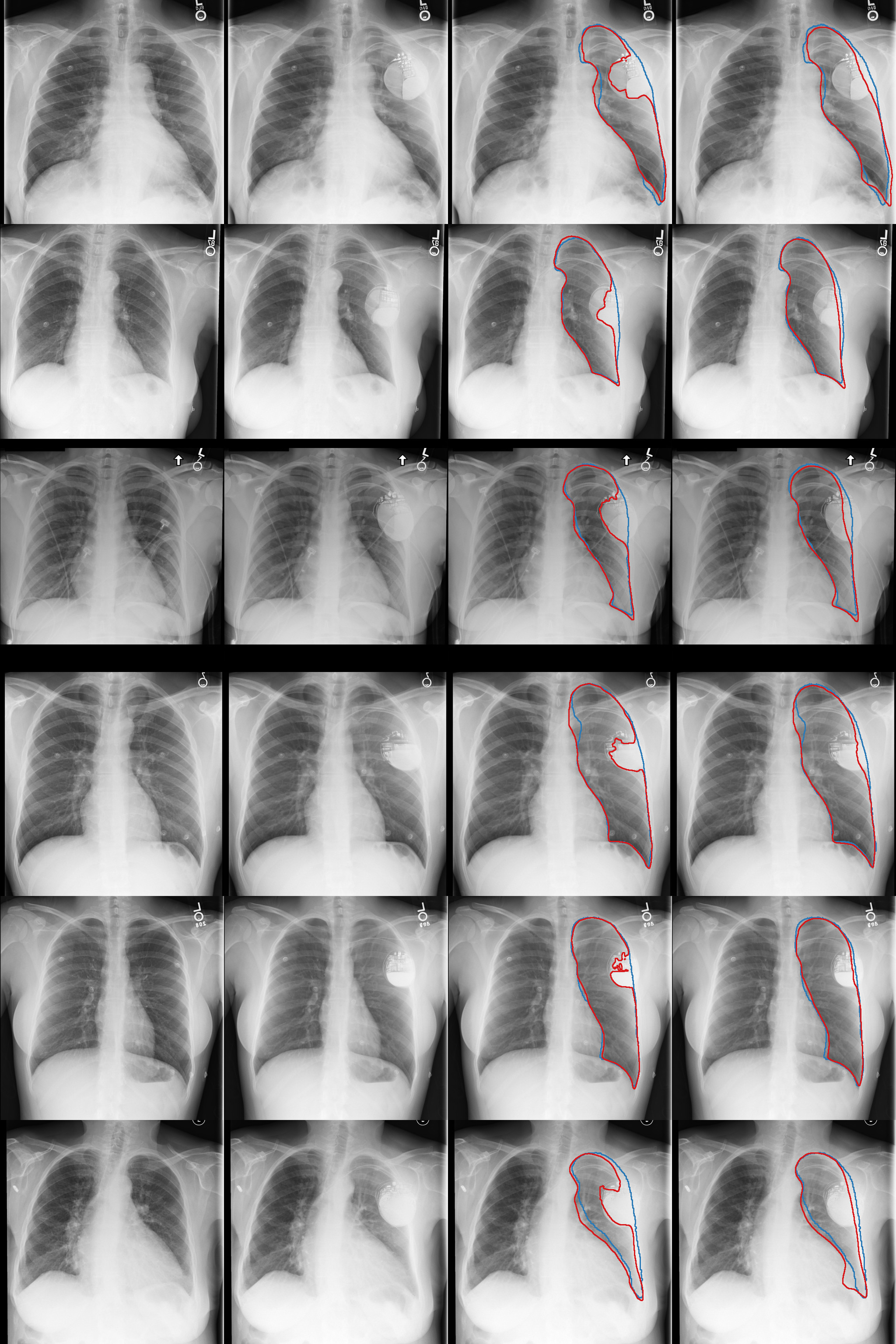}
        \hfill\subcaptionbox{\small{Original}}[0.24\linewidth]{\centering}
        \hfill\subcaptionbox{\small{Edited}}[0.24\linewidth]{\centering}
        \hfill\subcaptionbox{\small{Weak\\Predictor}}[0.24\linewidth]{\centering}
        \hfill\subcaptionbox{\small{Strong\\Predictor}}[0.24\linewidth]{\centering}
        \hfill\null
    \end{minipage}
    \caption{Additional edits simulated by \methodname for stress-testing two segmentation models. The `weak predictor' (c) and the `strong predictor' (d) are trained on MIMIC-Seg \cite{chen_chest_2022} and CheXmask \cite{gaggion_chexmask_2023} respectively, by adding pacemakers, which can be seen in the top left of images, via the prompt \prompt{Left pectoral pacemaker in place. The position of the leads is as expected. Otherwise unremarkable chest radiographic examination.} \textcolor{myblue}{Blue}: ground-truth mask: ; \textcolor{myred}{red}: predicted. Similar to the example in \cref{fig:segmentation-examples}, the segmentation model trained on MIMIC-Seg (which contains predominantly healthy patients) incorrectly segments around the pacemakers, while the model trained on CheXmask (which is larger and contains various abnormal cases), segments more accurately.}
    \label{fig:additional-pacemaker-edits}
\end{figure}

\begin{figure}[p]
    \centering
    \captionsetup[subfigure]{justification=centering, skip=-0.7em}
    \begin{minipage}{0.8\linewidth}
        \includegraphics[width=\linewidth]{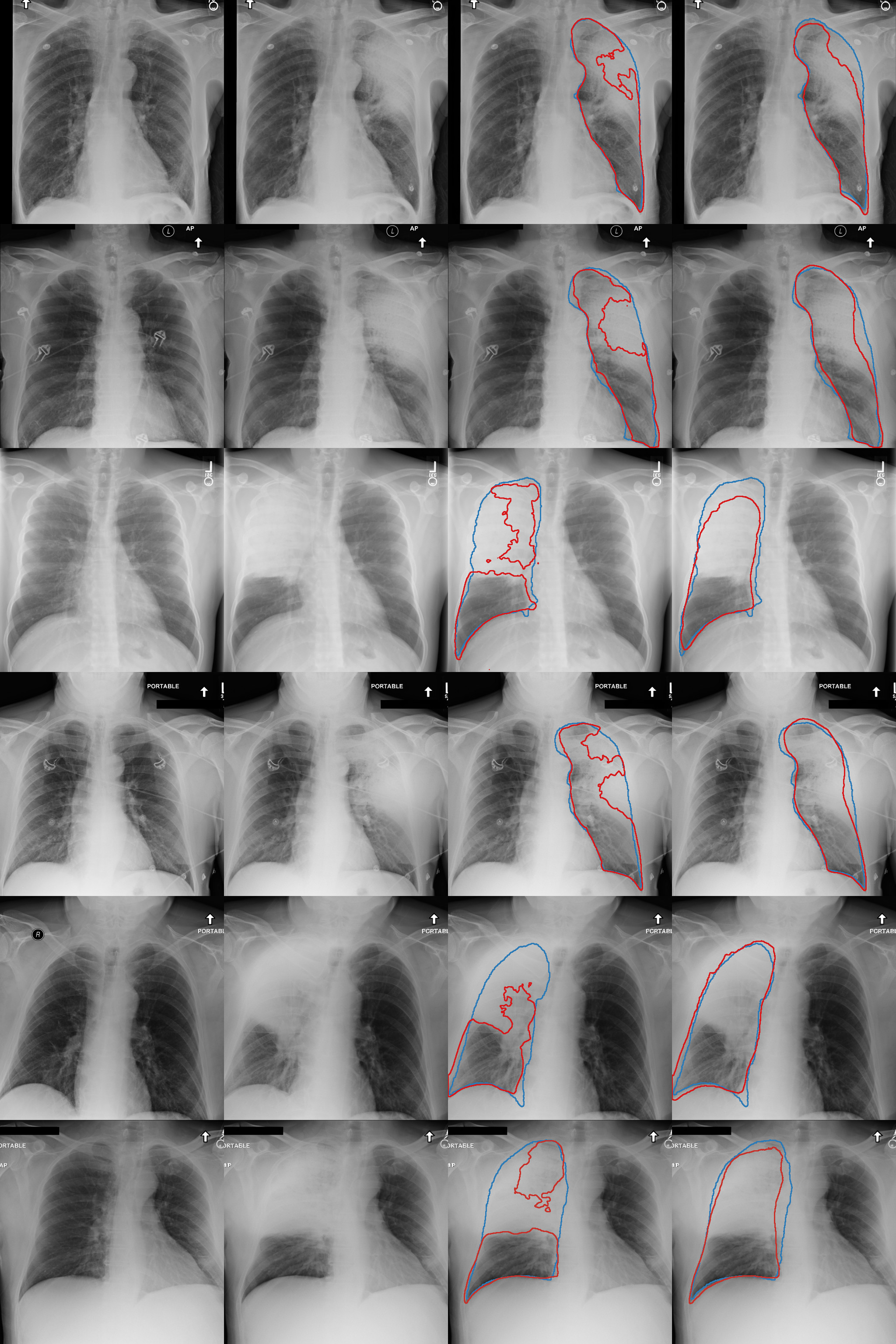}
        \hfill\subcaptionbox{\small{Original}}[0.24\linewidth]{\centering}
        \hfill\subcaptionbox{\small{Edited}}[0.24\linewidth]{\centering}
        \hfill\subcaptionbox{\small{Weak\\Predictor}}[0.24\linewidth]{\centering}
        \hfill\subcaptionbox{\small{Strong\\Predictor}}[0.24\linewidth]{\centering}
        \hfill\null
    \end{minipage}
    \caption{Additional edits simulated by \methodname for stress-testing two segmentation models. The `weak predictor' (c) and the `strong predictor' (d) are trained on MIMIC-Seg \cite{chen_chest_2022} and CheXmask \cite{gaggion_chexmask_2023} respectively, by adding upper-lobe consolidation, via the prompt \prompt{New [left/right] upper lobe consolidation.} \textcolor{myblue}{Blue}: ground-truth mask: ; \textcolor{myred}{red}: predicted. Similar to the example in \cref{fig:segmentation-examples}, both models are less able to segment the lungs accurately, however, segmentations by the model trained on MIMIC-Seg are notably worse, often excluding the consolidated region.}
    \label{fig:additional-consolidation-edits}
\end{figure}

In \Crefrange{fig:additional-edema-edits}{fig:additional-consolidation-edits} we show more examples of edits produced by \methodname to stress test the segmentation models. \methodname edits are high-quality, with both general anatomy maintained after the edit, as well as image markings.

\end{document}